\pdfoutput=1

\documentclass[11pt]{article}

\usepackage{acl}

\usepackage{times}
\usepackage{latexsym}

\usepackage[T1]{fontenc}

\usepackage[utf8]{inputenc}

\usepackage{microtype}

\usepackage{inconsolata}
\usepackage{mdframed} 
\usepackage{amsmath}
\usepackage{tabularx}
\usepackage{threeparttable}
\usepackage{booktabs}
\usepackage{amssymb}
\usepackage{multirow}
\usepackage{graphicx}
\usepackage{pgfplots}
\usepackage{pgf-pie} 
\usepackage{enumitem}
\usepackage{tcolorbox}
\usepackage{xcolor,colortbl}
\usepackage{subcaption}
\usepackage{tikz}
\usepackage{authblk}
\newcolumntype{P}[1]{>{\arraybackslash}p{#1}}
\newcolumntype{M}[1]{>{\centering\arraybackslash}m{#1}}

\usepgfplotslibrary{fillbetween} 
\pgfplotsset{compat=1.17} %

\usepackage{float}
\DeclareUnicodeCharacter{0307}{\'}

\renewcommand\AB@affilsepx{, \protect\Affilfont}

\newcommand{\up}{\textcolor{green!60!black}{$\uparrow$}}
\newcommand{\down}{\textcolor{red!60!black}{$\downarrow$}}

\newcommand{\samewhite}{\textcolor{white}{$\uparrow$}}
\newcommand{\rup}{\textcolor{red!60!black}{$\uparrow$}}
\newcommand{\gdown}{\textcolor{green!60!black}{$\downarrow$}}

\newcommand{\slike}{\textit{SL}}
\newcommand{\bse}{\textit{BSE}}
\newcommand{\cse}{\textit{CSE}}
\newcommand{\asc}{\textit{ASC}}
\newcommand{\psc}{\textit{PSC}}
\newcommand{\fsc}{\textit{PSC-F1}}

\title{Calibrating Long-form Generations from Large Language Models}

\author[1]{\textbf{Yukun Huang}}
\author[2]{\textbf{Yixin Liu}}
\author[1]{\textbf{Raghuveer Thirukovalluru}}
\author[2]{\authorcr\textbf{Arman Cohan}}
\author[1]{\textbf{Bhuwan Dhingra}}

\affil[1]{Duke University}
\makeatletter
\renewcommand\AB@affilsepx{\\ \protect\Affilfont}
\makeatother
\affil[2]{Yale University}

\makeatletter
\renewcommand\AB@affilsepx{, \protect\Affilfont}
\makeatother
\affil[ ]{\texttt{\{yukun.huang, raghuveer.thirukovalluru\}@duke.edu}}
\makeatletter
\renewcommand\AB@affilsepx{\\ \protect\Affilfont}
\makeatother
\affil[ ]{\texttt{bdhingra@cs.duke.edu}}
\affil[ ]{\texttt{\{yixin.liu, arman.cohan\}@yale.edu}}

\begin{document}
\maketitle
\begin{abstract}
To enhance Large Language Models' (LLMs) reliability, calibration is essential---the model's confidence scores should align with the likelihood of its responses being correct. However, traditional calibration methods typically rely on a binary true/false assessment of response correctness, unsuitable for long-form generations where an answer can be partially correct. 
Addressing this gap, we introduce a unified calibration framework, in which both the correctness of the LLMs' responses and their associated confidence levels are treated as distributions across a range of scores. 
We develop three metrics for assessing LLM calibration and propose confidence elicitation methods based on self-consistency and self-evaluation. 
Our experiments demonstrate that larger models don't necessarily guarantee better calibration, that various calibration metrics complement each other, and that self-consistency methods excel in factoid datasets. 
We also find that calibration can be enhanced through techniques such as fine-tuning, scaling the temperature. Finally, we illustrate one application of long-form calibration through selective answering in long-form responses, optimizing correctness within a constrained API budget.

\end{abstract}

\begin{figure}[h]
\includegraphics[width=\columnwidth]{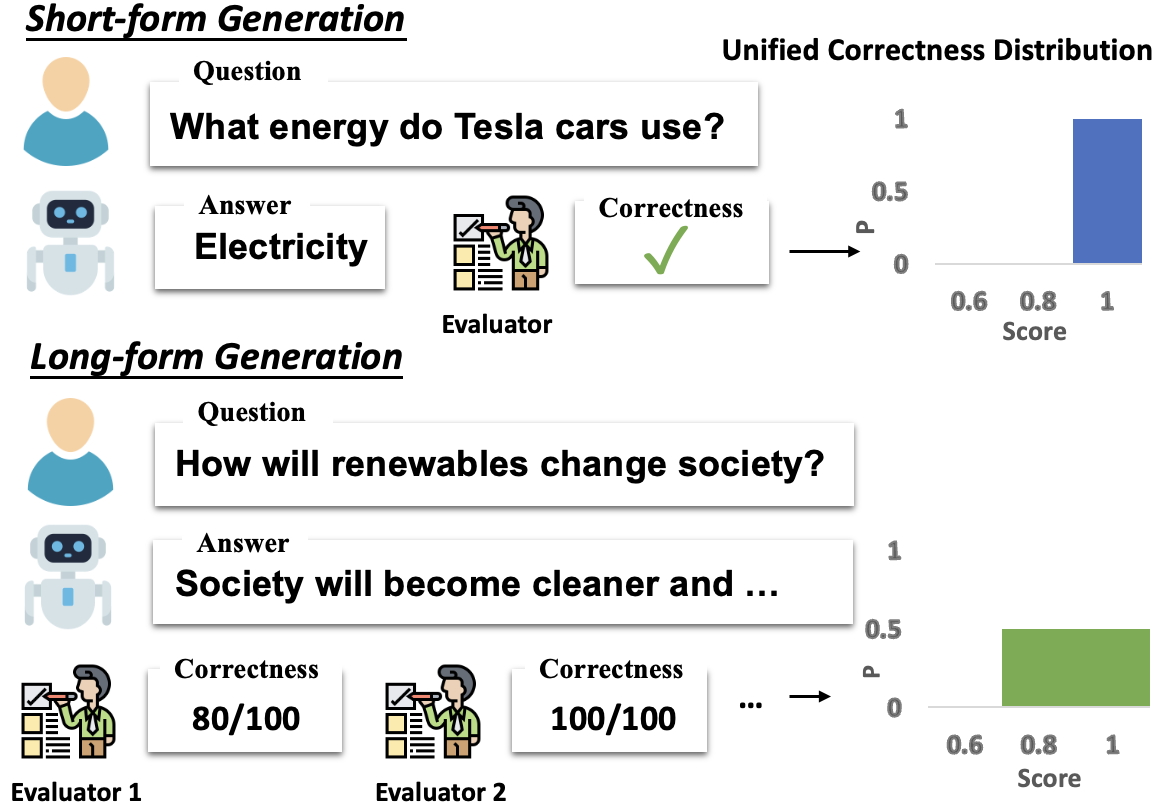}
\caption{A comparison between short-form generation and long-form generation. The correctness of the short-form answer can either be true (1) or false (0), while the correctness of the long-form answer is typically a score between 0 and 1. Both of these scores may vary across evaluators due to subjectivity, hence we conceptualize them as a distribution over $[0, 1]$.
}
\label{fig:comparison}
\end{figure}

\begin{figure*}[t]
\includegraphics[width=\textwidth]{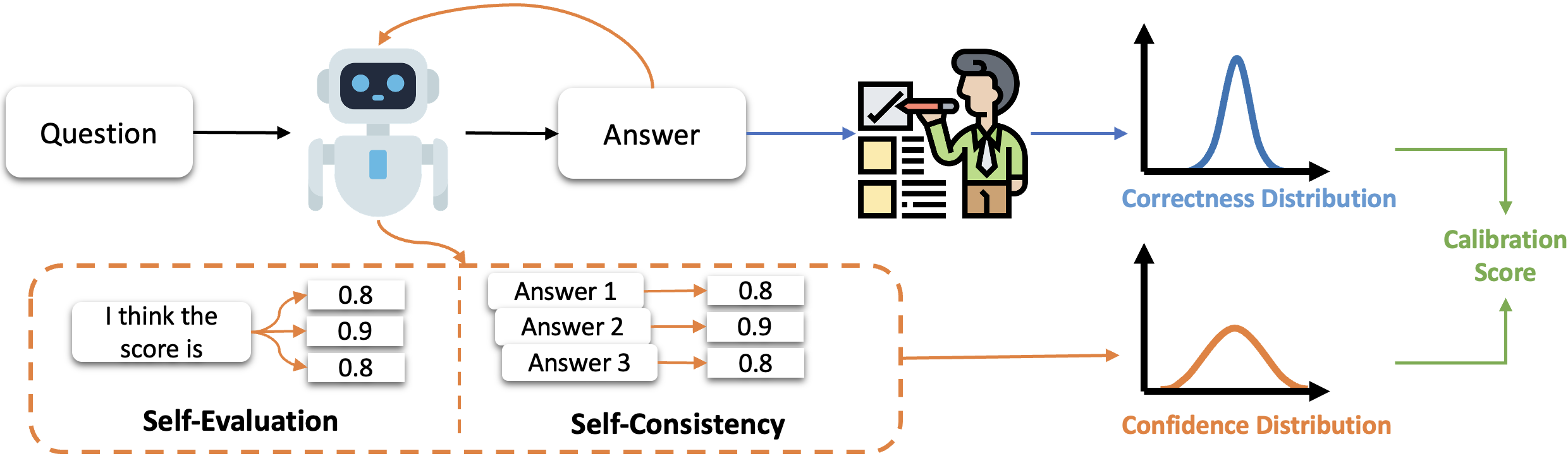}
\caption{Overview of our calibration framework. We prompt an LLM to produce an answer to a specific question, assess the answer's \textcolor{blue!70!black}{correctness distribution} using an evaluator (task-specific metric/GPT-4 metric/human metric), and determine the model's \textcolor{orange!70}{confidence distribution} through self-evaluation or self-consistency approaches. Finally, we calculate the \textcolor{green!60!black}{calibration score} by comparing the correctness and confidence distributions against our predefined metrics.}
\label{fig:system}
\end{figure*}

\section{Introduction}

Confidence calibration in large language models (LLMs) aims to align the model's internal confidence with a probabilistic perspective of its answers' correctness (i.e. quality), enhancing reliability and interpretability for aiding human decision-making \cite{Kadavath2022LanguageM}. People intuitively understand and utilize probabilities \cite{Cosmides1996AreHG}, making this approach crucial for practical applications.
Conventional calibration \cite{Guo2017OnCO} treats answer correctness as binary (true or false) and seeks to align the model’s confidence with the likelihood of model’s answer being correct, typically stated as: ``\textit{I am x\% confident that this answer is completely correct.}'' However, the correctness of long-form generation is not always either true or false but can be partially correct (Figure \ref{fig:comparison}). 

Therefore, a single confidence score for long-form outputs is ambiguous: it can either imply ``\textit{I am x\% confident that the answer is 100\% correct}'' or ``\textit{I am 100\% confident that the answer is x\% correct.}'' The former fails to capture the graded notion of long-form answer correctness, while the latter focuses on self-evaluation of correctness, rather than calibration as it overlooks the confidence at specific correctness levels.

Addressing this challenge, we propose to conceptualize the model's confidence as \textit{distribution} across scores between $[0, 1]$ to capture the nuanced understanding of the model of each correctness level of the long-form answer, corresponding to the statement \textit{``I am $x$\% confident that this answer is $y$\% correct''}. Moreover, we also view the correctness of an LLM's response as a \textit{distribution} across scores between $[0, 1]$ to capture graded and subjective assessments of long-form generations quality. This subjectivity arises from the multifaceted nature of evaluating long-form outputs, where factors like factuality, coherence, clarity, and comprehensiveness each play a role, potentially introducing variability in judgment~\cite{Bakker2022FinetuningLM}. We can then measure both the classical notion of calibration error, averaged across different correctness levels, as well as new notions of alignment between the correctness
and confidence distributions and their utility in selective prediction (\S~\ref{sec: calibration metric}).
\autoref{fig:system} shows an overview of our framework, which consists of three modular components: estimating the target correctness distributions, eliciting confidence distributions from LLMs, and measuring calibration between these distributions.

Our unified framework offers three key advantages. 
1. Generalizability: Our framework applies to both long-form and short-form generation tasks by representing correctness and confidence as distributions, regardless of whether the correctness of task is binary, continuous, subjective, or objective.
2. Flexibility: The framework is evaluation-metric agnostic, allowing the integration of any metric or confidence elicitation method, and can adapt as evaluation methods evolve.
3. Interpretability: It provides a nuanced view of uncertainty, enabling decision-makers to assess confidence across multiple correctness levels, fostering greater transparency and trust in the model’s outputs.

We leverage our framework to measure calibration for several LLMs on multiple datasets across three long-form QA---ASQA \citep{stelmakh-etal-2022-asqa}, ELI5 \citep{fan-etal-2019-eli5}, QAMPARI \citep{Amouyal2022QAMPARIAO}---and one summarization task, CNNDM~\citep{nallapati-etal-2016-abstractive}. Our results show that our methods excel over baselines by leveraging the model’s nuanced confidence distribution, stronger LLMs like GPT-3.5 don't necessarily guarantee better calibration, that various calibration metrics complement each other, and LLMs exhibit better calibration on factoid datasets than more open-ended datasets. 
Furthermore, our analysis highlights that fine-tuning and temperature scaling could enhance calibration. 
Finally, we illustrate a practical application of long-form calibration: employing a cascading strategy \citep{Chen2023FrugalGPTHT} for selective answering to optimize the cost-effectiveness of long-form text generation. In this approach, an open-source model initially handles queries and, based on its confidence levels---assessed using our system---a more advanced API model is engaged as needed. This method ensures cost efficiency while maintaining high-performance levels. 

In summary, our contributions are: 
\begin{itemize}
\setlength{\itemsep}{0pt}
    \item A universal calibration framework for text generation tasks, enhancing LLM evaluation for critical applications.
    \item  Innovative methods for confidence elicitation and calibration measurement, applied to a variety of LLMs.
    \item Evidence that calibration can be improved by model fine-tuning and temperature scaling.
    \item A cost-effective model usage strategy, illustrating the practicality of long-form calibration in optimizing LLM deployment.
\end{itemize}

\section{Related Work}
\textbf{Measuring Calibration} Calibration  (\citealp{Guo2017OnCO}, \citealp{Minderer2021RevisitingTC})has been widely studied in language models, whose probabilities derived from logits are generally found to not be calibrated (\citealp{jiang-etal-2020-know}, 
\citealp{Kadavath2022LanguageM}, \citealp{chen-etal-2023-close}). Standard metrics to measure the calibration include Expected Calibration Error (ECE) for confidence-accuracy disparity \citep{Naeini2015ObtainingWC}, Brier Score for mean squared prediction-outcome differences, and AUROC for assessing confidence-based correct answer identification (\citealp{Boyd2013AreaUT}, \citealp{Kuhn2023SemanticUL}). Selective Accuracy@Coverage measures accuracy within the model's most confident predictions (\citealp{Liang2023HolisticEO}, \citealp{cole-etal-2023-selectively}).  However, these metrics, rooted in a binary notion of correctness, fall short for long-form tasks where correctness spans a range, suggesting a distribution-based approach is more apt.
\\ \\
\textbf{Improving Calibration} Traditional calibration methods focus on post-processing logits~\citep{Guo2017OnCO}, but with LLMs generating unbounded text, logits could fall short. Thus, extracting better confidence scores (i.e., confidence elicitation) has become crucial for improving calibration. These include:  \textit{verbalization}, which directly asks the model to output its confidence (\citealp{Lin2022TeachingMT}), \textit{consistency}, which uses the uniformity of multiple responses to gauge confidence (\citealp{Kadavath2022LanguageM}, \citealp{Kuhn2023SemanticUL}, \citealp{cole-etal-2023-selectively}, \citealp{chen-etal-2023-relation}, \citealp{Tian2023FinetuningLM}, \citealp{Lin2023GeneratingWC}), and the \textit{hybrid} of both (\citealp{Xiong2023CanLE}, \citealp{tian-etal-2023-just}, \citealp{Chen2023QuantifyingUI}).  However, these methods often presume binary answer correctness, offering a singular confidence score that fails to capture the nuanced correctness required for long-form tasks.  Recent work \citep{Zhang2024LUQLU} on long-form generation addresses continuous correctness scores but focuses on aligning an uncertainty score with correctness, rather than on improving calibration.

\section{Long-form Generations Calibration}
This section formalizes the long-form generation calibration problem (\S~\ref{subsection: formulation}), and introduces three core components in our calibration framework (\autoref{fig:system}): the correctness distributions of the answers (\S~\ref{subsection: target correctness}), the confidence distributions of the LLM on its answers (\S~\ref{sec: confidence elicitation}), and the calibration metrics to measure how well these two align (\S~\ref{sec: calibration metric}). 

\subsection{Formulation}
\label{subsection: formulation}
Given a dataset $\mathcal{D}$, the model's answer for each question $Q_i$ in the dataset is answer $A_i$ (where $i$ indexes the questions in the dataset). To measure how calibrated the model is, we need three steps. 
First, we apply an evaluator to get the target correctness distribution $P_{T_i}$ where $T_i$ is the random variable that denotes the correctness score in answer $A_i$. 
\vspace{-5pt}
\begin{equation}
    P_{T_i}(x) = \text{Pr}(A_i \text{ is } s \text{ correct})
\end{equation} for $s\in \mathcal{S}$, where $\mathcal{S}$ is the space of correctness levels (e.g., normalized from ordinal scores ranging from 0 to 5). It should be noted that $s$ could theoretically be a continuous value in the range $[0\%, 100\%]$. However, since humans tend to make more accurate judgments using discrete ratings due to “rounding bias”~\citep{Honda2022OnTR}, we approximate continuous correctness with ordinal scores in practice. Second, We use a confidence elicitation method to derive the confidence distribution $P_{C_i}$ from LLM $\mathcal{M}$, where $C_i$ represents the model's confidence in its answer $A_i$. We ensure these confidence scores are normalized to form a valid distribution, matching the domain of the correctness distribution. For $s\in \mathcal{S}$,
\begin{align}
    P_{C_i}(x) = \text{$\mathcal{M}$'s confidence that \textit{A} is \textit{s} correct}
\end{align}  
Finally, we design metrics to measure the alignment between $P_{C_i}$ and $P_{T_i}$ across the dataset.

\subsection{Correctness Distribution Estimation}
\label{subsection: target correctness}
To establish correctness distributions as alignment targets, we need to adopt long-form evaluation metrics that integrate aspects such as relevance, coherence, factuality, and helpfulness. Traditional metrics like BLEU and ROUGE fail to capture semantic meaning \cite{liu-etal-2023-g}, while factuality-based metrics like FactScore \cite{min-etal-2023-factscore} may neglect question relevance. 
GPT-4 metrics have gained popularity \cite{Li2024LeveragingLL} due to their adaptability and comparative accuracy (\citealp{jain-etal-2023-multi}, \citealp{liu-etal-2023-g}). These metrics allow for the integration of various user-prioritized aspects by adjusting evaluation rubrics, providing a balanced approach to both referenced and divergent answers. However, they also have limitations like a bias toward longer outputs \cite{Zheng2023JudgingLW}. 

Our framework is evaluator-agnostic, allowing us to use any correctness distribution metric. To identify a practical and actionable metric that best guides human decision-making, we conduct human evaluations to determine alignment with human judgment. According to the results in Appendix \ref{appendix: evaluation metric}, we determined that the GPT-4 metric, with a higher correlation to human judgments compared to the task-specific metric, is more effective for three datasets (e.g., 76.2 v.s 47.8 in ASQA), while the task-specific metric is preferred for another dataset. Our framework’s modular design allows replacements of evaluation metrics based on user needs or new developments.

\subsection{Confidence Distribution Elicitation}
\label{sec: confidence elicitation}
There are two common strategies to develop confidence in model responses: explicitly asking the model to verbalize its confidence or implicitly estimating it through self-consistency. However, the single confidence score provided by prior studies is ambiguous and non-interpretable in long-form calibration. 
Therefore, we develop two methods tailored for long-form calibration accordingly.
\\
\textbf{Continuous Self Evaluation} (\cse)
We prompt the model to repeatedly perform self-evaluations, where the resulting scores (typically ordinal scores from 0 to 5) are normalized to a [0,1] range and interpreted as a confidence distribution. The self-evaluation template closely mirrors the template for correctness evaluation but omits the reference answer (see Appendix \ref{subappendix: self-evaluation template} for details). Formally, given an LLM $\mathcal{M}$, $N$ self-evaluations: 
\begin{align}
    P_{C_i}(s) &= \frac{1}{N}\sum_{j=1}^N \mathrm{1}(\mathcal{M}(A_i)_j = s)
\end{align}
for score $s\in \mathcal{S}$, where $\mathcal{S}$ is the space of correctness levels.
Such a sampling method provides a more authentic reflection of the model's internal distribution than logits~\citep{cole-etal-2023-selectively}. By asking the model to assess an answer multiple times, we capture a range of scores that better represent the model's confidence, which enhances the reliability of the confidence estimation. 
\\
\textbf{Pairwise Self Consistency} (\psc)
Another key indicator of model confidence is the consistency among multiple responses a model provides for a given question. Given a primary answer $A_i$, other $N$ answers $A_i^1 ... A_i^N$ sampled from LLM $\mathcal{M}$, a metric for measuring the similarity between two answers $Sim(\cdot, \cdot)$, and a score $s\in \mathcal{S}$:
\vspace{-10pt}
\begin{align}
    P_{C_i}(s) &= \frac{1}{N}\sum_{j=1}^N \mathrm{1}(Sim(A_i, A_i^j) = s)
\end{align}
Assessing similarity in long-form answers is more complex than with short responses \citep{Kadavath2022LanguageM}. To address this, we propose four methods for evaluating similarity in long-form content: 1. \textbf{Naive}, assessing overall response similarity with an LLM; 2. \textbf{Split}, analyzing sentence-level similarity; 3. \textbf{Claim}, evaluating claim matching; 4. \textbf{Named Entity Recognition} (NER), focusing on named entity overlap. 
These approaches range from broad to detailed analysis, chosen based on task requirements and the desired analysis depth. See Appendix \ref{appendix: self-consistency} for more details.

\subsection{Calibration Metrics}
\label{sec: calibration metric}
We introduce three key metrics to assess model calibration from various angles. \textit{Expected correctness error with multi-class} (ECE-M) measures the alignment between the model’s stated confidence in reaching a particular level of correctness and the actual likelihood that the model performs at that specified level, across the spectrum from 0 to 1. \textit{Correlation} evaluates the alignment between expected confidence and correctness \textit{across the dataset}, indicating the model's proficiency in ranking answers. \textit{Selective F1} measures the utility of confidence scores in identifying the good answers and abstaining from the rest.
\\ \\
\textbf{ECE-M}
The classical notion of calibration relies on an (answer-correctness) pair of random variables $(A, Y)$ $\in \mathcal{A} \times \{0,1\}$, where $\mathcal{A}$ is the answer space. An LLM $\mathcal{M}$ with its confidence elicitation method $h_\mathcal{M}$: $\mathcal{A} \rightarrow [0, 1]$ is said to be well-calibrated if $\text{Pr}(Y=1|h_\mathcal{M}(A)=q)=q$ for $q\in[0,1]$. 
To measure if this holds, traditional ECE(h) \cite{Gupta2021ToplabelCA} is: 
$\mathbb{E}_A\left[\left|\text{Pr}(Y=1|h_\mathcal{M}(A))-h_\mathcal{M}(A)\right|\right]$

In long-form calibration where the answer correctness is a continuum $Y \in [0, 1]$, the probabilistic confidence predictor $h_\mathcal{M}$ should predict confidence about each level of correctness and therefore denoted as $h_\mathcal{M}$: $\mathcal{A} \times [0,1] \rightarrow [0, 1]$. The notion of long-form calibration is $\text{Pr}(Y=s|h_\mathcal{M}(A, s)=q_s)=q_s$ for every $s\in[0,1]$ and $q_s\in[0, 1]$
We define ECE-M as the aggregation of ECE scores for all correctness levels. In practice, we use discrete levels $s \in \mathcal{S}$
(e.g., ratings from 0-5) for the correctness scores.
Hence,
we calculate an ECE$(s,h)$ conditioned on each $s$, 
\begin{align}
    \mathbb{E}_X\left[\left|\text{Pr}(Y=s|h_\mathcal{M}(A, s))-h_\mathcal{M}(A,s)\right|\right]
\end{align}

Then the final ECE-M score is weighted by the frequency of each class: 
\begin{align}
    \text{ECE-M}(h):=\sum_{s\in \mathcal{S}}\text{Pr}(Y=s)\text{ECE}(s,h)
\end{align}
\vspace{-20pt}
\\ \\
\textbf{Correlation}
While ECE-M focuses on measuring calibration at each correctness level independently, it doesn’t account for the distance between these levels. To address this, we also measure the correlation between the expected values of the confidence and correctness distributions for a more comprehensive assessment. For each answer $A_i$ within the dataset, we can calculate the expected correctness score of the model $E|C_i|=\sum_sP_{C_i}(s)\times s$ and the expected correctness score of the 
 target $E|T_i|=\sum_sP_{T_i}(s)\times s$. In the whole dataset, we can get a list of expected correctness scores $\mathbf{E_C}$ from model confidence and a list of target expected correctness scores $\mathbf{E_T}$. Then we can measure the correlation between them. 
\begin{align}
    \rho(\mathcal{D})=Corr(\mathbf{E_C}, \mathbf{E_T})
\end{align}
where $Corr(\cdot)$ represents the correlation function. This correlation provides a clear indicator of how well our model's confidence aligns with its actual correctness across the entire dataset.
\\ \\
\textbf{Selective F1}
In selective answering (e.g., \citealp{kamath-etal-2020-selective}), models only respond when confident about their accuracy to improve reliability. Traditional metrics for selective answering include accuracy@coverage and coverage@accuracy \citep{tian-etal-2023-just}, 
which measures a model's precision and recall in selecting completely correct answers. Similarly, in long-form calibration, it is crucial to assess the model's selection of answers that are at least "s\% correct" using both precision and recall. 

Therefore,  we propose the selective F1 metric ($F1_{\tau_s}$) to quantify the model's aptitude in filtering out answers that meet or exceed a predefined correctness threshold $\tau_s$. Our approach utilizes a dual-threshold system, consisting of a confidence threshold ($\tau_c$) and the correctness threshold ($\tau_s$), allowing the model to answer questions only if its confidence in the answer's expected correctness score exceeding $\tau_s$ surpasses $\tau_c$. Formally, let $\mathcal{A}=\{A_1, A_2, ..., A_n\}$ denote the total set of model's answers in the dataset and $\mathcal{A}^*=\{A_i \in \mathcal{A} \mid  \sum_{s\geq \tau_s} P_{C_i}(s) \geq \tau_c \}$ denote the set of selected answers. 
Let the indicator function $I_{\tau_s}(A_i)$ indicate if the expected correctness score $E[T_i]$ of $A_i$ exceeds $\tau_s$:
\vspace{-10pt}
\begin{align}
    I_{\tau_s}(A_i) = 
    \begin{cases}
        1 & \text{if } E[T_i] \geq \tau_s \\
        0 & \text{Otherwise}
    \end{cases}
\end{align}
The selective precision $P_{\tau_s}$ on the dataset $\mathcal{D}$ is the proportion of selected answers that surpass the correctness threshold $\tau_s$ relative to the total number of selected answers:
\begin{align}
    P_{\tau_s}(\mathcal{D}) = \frac{\sum_{A_i \in \mathcal{A}^*} I_{\tau_s}(A_i)}{|\mathcal{A}^*|}
\end{align}
The selective recall $R_{\tau_s}$ compares the number of selected answers meeting this criterion against the total number of correct answers in the dataset that exceed the threshold $\tau_s$:
\begin{align}
    R_{\tau_s}(\mathcal{D}) = \frac{\sum_{A_i \in \mathcal{A}^*} I_{\tau_s}(A_i)}{\sum_{A_i \in\mathcal{A}} I_{\tau_s}(A_i)}
\end{align}
The selective F1 combines recall and precision: 
\begin{align}
    F1_{\tau_s}(\mathcal{D}) = 2 \frac{P_{\tau_s}(\mathcal{D})R_{\tau_s}(\mathcal{D})}{P_{\tau_s}(\mathcal{D}) + R_{\tau_s}(\mathcal{D})}
\end{align}
In our experiments, we select  $\tau_s$  as the nearest correctness level greater than the best LM’s average correctness score. For $\tau_c$, we choose the value that yields the highest selective F1 score on the development split.
\setlength{\tabcolsep}{4pt}
\begin{table*}[t!]
\small
\centering
\begin{tabular}{l|ccc|ccc|ccc|ccc}
\toprule
& \multicolumn{3}{c|}{\cellcolor{gray!15}\textbf{ASQA}} & \multicolumn{3}{c|}{\cellcolor{gray!15}\textbf{QAMPARI}} & \multicolumn{3}{c|}{\cellcolor{gray!15}\textbf{ELI5}} & \multicolumn{3}{c}{\cellcolor{gray!15}\textbf{CNNDM}}\\
Method & ECE-M & Corr & $F1_{0.8}$ & ECE-M & Corr & $F1_{0.4}$ & ECE-M & Corr & $F1_{0.8}$ & ECE-M & Corr & $F1_{0.8}$ \\ 
\midrule
SL$^*$ & 28.2\samewhite & 0.7\samewhite & 0.0 \samewhite & 27.4\samewhite & 10.9\samewhite & 8.0\samewhite  & 29.6\samewhite & -11.9\samewhite & 0.0\samewhite & 77.2\samewhite & -7.5\samewhite & 0.0\samewhite\\
\midrule
BSE$^*$ & 32.8\samewhite & 14.2\samewhite & 57.6 \samewhite & \textbf{25.2}\samewhite & 16.8\samewhite & 33.2\samewhite  & 30.3\samewhite & 11.7\samewhite & 46.7\samewhite & 78.5\samewhite & 11.2\samewhite & 90.6\samewhite\\
CSE & 29.0\gdown & 16.3\up & 58.5 \up & 42.8\rup & 21.9\up & 33.4\up  & 31.2\rup & \textbf{26.9}\up & \textbf{48.2}\up & \textbf{15.2}\gdown & \textbf{19.2}\up & \textbf{92.0}\up\\
\midrule
ASC$^*$ & 35.9\samewhite & 27.1\samewhite & 5.2 \samewhite & 46.0\samewhite & 38.6\samewhite & 38.5\samewhite  & 38.4\samewhite & 16.7\samewhite & 7.9\samewhite & 63.2\samewhite & 8.8 \samewhite & 44.2\samewhite\\
PSC$_{F1}$ & 28.8\samewhite & 27.1\samewhite & 33.5 \samewhite & 38.4\samewhite & 38.6\samewhite & 42.7\samewhite  & 27.1\samewhite & 16.7\samewhite & 20.7\samewhite & 57.1\samewhite & 8.8\samewhite & 79.5\samewhite\\
PSC & \textbf{18.3}\gdown & \textbf{46.8}\up & \textbf{61.6} \up & 26.2\gdown & \textbf{39.1}\up & \textbf{44.0}\up  & \textbf{24.9}\gdown & 24.9\up & 46.2\up & 64.5\rup & 15.5\up & 90.0\up\\
\bottomrule

\end{tabular}

\caption{Calibration Performance Comparison Among Different Confidence Elicitation Methods Across Four Tasks (in \%): ``ECE-M'' for expected correctness error with multi-class, ``Corr'' for Correlation,  ``$F1_{\tau_s}$'' for Selective F1 Score at threshold $\tau_s$. Results represent averages from five models. Methods with * served as baselines. For ``Corr'', ``$F1_{\tau_s}$'', and ``Score'', \up{} means better than corresponding baseline while \down{} is worse. For ``ECE-M'', \gdown{} is better while \rup{} is worse. The best score among all confidence elicitation methods is \textbf{bolded}. Key insights: 1) Self-Consistency (PSC) outperforms Self-Evaluation on factoid datasets; 2) Our methods PSC and CSE surpass baselines; 3) Different metrics offer complementary insights
}
\label{table:main_results}
\end{table*}

\section{Experiments and Results}
\subsection{Setup}
\textbf{Models and Data} We measure different sized LLMs' calibration, including Llama-2-13b-chat, Llama-2-70b-chat \citep{Touvron2023Llama2O}, Vicuna-13b \citep{Zheng2023JudgingLW}, Llama-3-8b-Instruct, GPT-3.5-turbo, across three long-form QA tasks: ASQA \citep{stelmakh-etal-2022-asqa}, ELI5 \citep{fan-etal-2019-eli5}, QAMPARI \citep{Amouyal2022QAMPARIAO}, and one summarization task: CNNDM \citep{nallapati-etal-2016-abstractive}. Details of datasets can be found in Appendix \ref{appendix:dataset}. 

\noindent\textbf{Correctness Evaluation} We apply GPT-4 to evaluate target correctness distributions for ASQA, ELI5, and CNNDM. In QAMPARI where the answer is a list of entities, we evaluate using the F1-5 metric, calculating the F1 score by the exact match with the gold answer and defining 100\% recall for predictions with at least 5 correct answers.

\subsection{Confidence Elicitation Methods}
In addition to our methods \cse{} and \psc{} (see Appendix \ref{Appendix: confidence elicitation} for similarity measurement choosing), we established baselines for self-evaluation, self-consistency, and logits-based approaches. This is because prior studies lack directly applicable baselines, primarily due to the non-interpretable nature of single confidence scores. 
\\
\textbf{Sentence Likelihood} (\slike): Based on prior studies using logits to gauge model confidence, we adopt sentence likelihood as a baseline measure, which typically results in a confidence distribution focused at the lowest score in long-form answers.
\\ 
\textbf{Binary Self-Evaluation} (\bse): following previous work \cite{Kadavath2022LanguageM} that asks model to self-evaluate if its answer is true several times, using the frequency of true as model's confidence score towards the answer being true. Then we adapt such a single score as a distribution focus solely on the values 0 and 1.
\\
\textbf{Average Self-Consistency} (\asc): Following prior work \cite{Xiong2023CanLE} using the average consistency between these candidate responses and the original answer then serves as a single measure of confidence score, we adopt the simple f1-token score to measure the consistency to adapt it to long-form generations. Then we treat the single score as a point mass distribution.
\\ 
\textbf{Pairwise Self-Consistency F1} (\fsc): Still using F1 to measure the consistency like \asc, but we directly treat the pairwise consistency scores as a distribution without aggregating, thereby keeping model intrinsic understanding about different correctness levels.

\subsection{Main Results}
In Table \ref{table:main_results}, we evaluate the calibration performances of various confidence elicitation methods by averaging the scores across all models.
Table \ref{table:main_results} shows that our methods, \cse{} and \psc{}, generally outperform their respective baseline categories and also surpass the logits-based method \slike. Key findings from the results include:
\\ \\
\textbf{Self-Consistency Outperforms Self-Evaluation on Factoid Datasets} 
Self-consistency methods typically outperform self-evaluation on factoid datasets like ASQA and QAMPARI. However, their effectiveness diminishes in more subjective tasks such as ELI5 or CNNDM. We hypothesize that this is because self-consistency is more readily quantifiable in factoid datasets, where the agreement between answers can be assessed based on factual consistency, thus providing clearer criteria. Conversely, in open-ended datasets, the consistency between answers is more ambiguous, making it more difficult to measure. \\ \\
\textbf{Nuanced Self-Evaluation Enhances Calibration}
\cse{} generally outperforms \bse{} by providing detailed confidence estimates at each correctness level.  However, the overall improvement remains constrained by the intrinsic limitations of LMs in self-evaluating their correctness, which sometimes hampers accurate estimations.
\\ \\
\textbf{Pair-wise Similarities Distribution and Task-tailored Similarity Measurement Help Calibration}
Both \asc{} and \fsc{} measure similarity with token-level F1. However, \fsc{} treats these scores as a distribution rather than averaging them, leading to better ECE-M and selective F1. \psc{} further enhances calibration by adopting a task-specific, detailed measurement of similarity, outperforming FSC in all four tasks.
\\
\begin{table*}[htbp]
\centering
\begin{tabular}{l|cccc|cccc}
\toprule
& \multicolumn{4}{c|}{\cellcolor{gray!15}\textbf{ASQA}} & \multicolumn{4}{c}{\cellcolor{gray!15}\textbf{QAMPARI}}\\
Model & ECE-M & Corr & $F1_{0.8}$ & Score & ECE-M & Corr & $F1_{0.4}$ & Score \\ 
\midrule
Llama2-13b & 15.9 & 48.0 & 47.5 & 51.3  & 30.5 & 46.0 & 42.3 &13.3 \\
Llama2-70b & 14.7 & 44.3 & 61.9 & 59.4 & 29.3 & 17.1 & 37.0 & 14.6 \\
Vicuna-13b & 20.1 & \textbf{58.2} & 56.8 & 50.8  & \textbf{14.4} & \textbf{49.5} & 42.6 & 11.4 \\
Llama-3-8b & \textbf{14.3} & 53.2 & 65.1 & 54.9  & 33.0 & 38.3 & 42.0 & 14.1 \\
GPT-3.5-turbo & 26.7 & 30.5 & \textbf{76.7} & \textbf{72.6}  & 23.7 & 44.4 & \textbf{56.2} & \textbf{24.0} \\
\end{tabular}

\begin{tabular}{l|cccc|cccc}
\toprule
& \multicolumn{4}{c|}{\cellcolor{gray!15}\textbf{ELI5}} & \multicolumn{4}{c}{\cellcolor{gray!15}\textbf{CNNDM}}\\
Model & ECE-M & Corr & $F1_{0.8}$ & Score & ECE-M & Corr & $F1_{0.8}$ & Score \\ 
\midrule
Llama2-13b & 36.0 & 21.8 & 40.1 & 53.8 & 12.6 & 19.5 & 92.0 & 77.0\\
Llama2-70b &  32.8 & 18.8 & 54.4 &61.7 & 13.8 & 6.4 & 93.6 & 77.6\\
Vicuna-13b  & \textbf{19.4} & 31.3 & 34.9 & 53.0 & 32.7 & 3.9 & 64.2 & \textbf{78.2}\\
Llama-3-8b  & 34.9  & \textbf{36.7} & 48.4 & 57.1 & 9.3 & \textbf{49.0} & 86.9 & 77.8\\
GPT-3.5-turbo & 32.7 & 26.2 & \textbf{63.4} & \textbf{63.0} & \textbf{7.8} & 17.0 & \textbf{94.7} & \textbf{78.2}\\

\bottomrule

\end{tabular}

\caption{Comparison of Calibration Performance Across Models for Four Tasks (in \%): We identify the optimal confidence elicitation method for each task and compare the performance of various models using this method. ``Score'' means the model's average correctness score on that task.  A key observation is that more powerful LMs do not necessarily exhibit better calibration, although they tend to perform better in selective answering.
}
\label{table:main_results_model}
\end{table*}
\\
\textbf{Calibration metrics complement each other}
A simplistic approach like \slike, which allocates all the probability mass to the point of score 0, can misleadingly show decent ECE-M (28.2\%) in specific cases like ASQA. However, its negative correlation (0.7\%) and zero F1$_{0.8}$ underscore an ineffective confidence distribution. Similarly, BSE in CNNDM may achieve a high F1$_{0.8}$ (90.6\%) by overestimating answer correctness, but this does not truly reflect response quality (correlation: 11.2\%) or provide well-calibrated probabilities across correctness levels, resulting in a bad ECE-M (ECE-M: 78.5\%). Hence, a comprehensive evaluation using multiple metrics is essential for a balanced assessment of model calibration.
\\ \\
\textbf{Larger models are not necessarily better calibrated.} In Table \ref{table:main_results_model}, we focus on the calibration performance of individual models when paired with the best-performing confidence elicitation method for each task.
Table \ref{table:main_results_model} shows that despite poor performance on ASQA and QAMPARI, Vicuna-13b has the highest correlation across these datasets. 
It might be because reinforcement learning for other models causes miscalibration by encouraging overfitting to rewarded behaviors \citep{Kadavath2022LanguageM}.
Scaling the temperature could enhance the calibration of LLMs fine-tuned using RL (see \S~\ref{subsec: improving calibration}). Additionally, Llama-2-13b demonstrates a higher correlation than its larger counterpart, Llama-2-70b. However, GPT-3.5-turbo, the strongest model, consistently scores the highest in selective F1 across all datasets. This performance can be attributed to the model’s ability to generate a larger volume of high-quality answers, increasing the probability of selecting superior responses even if it is not particularly well-calibrated. Consequently, the Selective F1 metric blends performance and calibration, and tends to favor more capable models due to their higher output of quality answers.
\\ \\

\subsection{Improving Calibration}
\label{subsec: improving calibration}
We delve into different strategies to enhance calibration: fine-tuning,  scaling the temperature, adding source documents (Appendix \ref{appendix: source documents}), and hybrid confidence elicitation (Appendix \ref{appendix: hybrid confidence elicitation}).  
\\ \\
\textbf{Fine-tuning}
Our study explores three fine-tuning strategies to improve model calibration on the ASQA dataset: fine-tuning the model for self-evaluation (using questions and model answers to produce scores and explanations), fine-tuning the model for generation (generating answers from questions), and a hybrid of both. GPT-4 synthesizes self-evaluation data by assessing different models' answers to questions from the ASQA training set, while the generation data originates from the ASQA training set itself. We apply LoRA \citep{Hu2021LoRALA} fine-tuning to the Llama2-13b model. See Appendix \ref{appendix:fine-tuning} for experiment details. As Table \ref{tab:fine-tuning} reveals, solely training on self-evaluation (`Evaluation') did not yield consistent improvements in calibration, possibly due to the complexity of this task and the limitation of LORA. Nonetheless, fine-tuning the model improves the self-consistency method, especially when the generation data is included during training (`Eval + Gen' and `Generation'). The model becomes more confident in terms of self-consistency after fine-tuning.

\begin{table}[h!]
\centering
\small
\begin{tabular}{lcccc}
\toprule
\textbf{Data} & \textbf{Corr} & \textbf{ECE-M} & \textbf{F1$_{0.8}$} & \textbf{Score} \\ 
\midrule
\multicolumn{5}{c}{\cellcolor{gray!15}\textbf{Self-Evaluation (CSE)}} \\
None & 18.1 \samewhite & 30.3 \samewhite & 50.8 \samewhite & 51.3 \samewhite \\ 
Evaluation & 13.6 \down & 32.4 \rup & 52.3 \up & 49.2 \down \\
Generation & 20.0 \up & 26.2 \gdown & 53.4 \up & 52.1 \up\\ 
Eval + Gen & 23.9 \up & 20.2 \gdown & 46.6 \down & 50.1 \down \\ 
\midrule
\multicolumn{5}{c}{\cellcolor{gray!15}\textbf{Self-Consistency (PSC)}} \\
None & 48.0 \samewhite & 15.9 \samewhite & 47.5 \samewhite & 51.3 \samewhite \\ 
Evaluation & 46.9 \down & 13.6 \gdown & 56.1 \up  & 49.2 \down \\
Generation & 58.9 \up & 14.5 \gdown & 59.5 \up & 52.1 \up\\ 
Eval + Gen & 54.5 \up & 12.2 \gdown & 50.7 \up & 50.1 \down \\ 
\bottomrule
\end{tabular}
\caption{Comparison among raw and fine-tuned Llama-2-13b on ASQA. ``None'' for the untrained model, ``Evaluation'' for the model fine-tuned with the self-evaluation dataset, ``Generation'' for the model fine-tuned with the ASQA generation data, and ``Eval+Gen'' for the model fine-tuned with the hybrid dataset combined by self-evaluation dataset and generation data.}
\label{tab:fine-tuning}
\end{table}

\begin{figure}[htb]
\centering
\begin{subfigure}[b]{0.23\textwidth}
\begin{tikzpicture}[scale=1]
\small
\begin{axis}[
    width=4cm,
    height=3cm,
    xlabel={$t$ },
    ylabel={Correlation(\%)},
    xmin=0.2, xmax=1,
    ymin=20, ymax=60,
    xtick={0.2, 0.4,0.6,0.8,1.0},
    ytick={20,40,60},
    legend pos=north west,
    legend style={at={(0.5,-0.55)},
      anchor=north, font=\tiny, legend columns=-1},
    ymajorgrids=true,
    grid style=dashed,
]
\addplot[
    color=green!60!black,
    mark=square,
    ]
    coordinates {
    (0.2, 25.0)  (0.4, 35.7) (0.6, 46.0)  (0.8, 47.8)  (1.0, 51.1)
    };
\end{axis}
\end{tikzpicture}
\end{subfigure}
\hfill
\begin{subfigure}{0.23\textwidth}
\begin{tikzpicture}[scale=1]
\small
\begin{axis}[
    width=4cm,
    height=3cm,
    xlabel={$t$},
    ylabel={ECE-M(\%)},
    xmin=0.2, xmax=1,
    ymin=10, ymax=50,
    xtick={0.2, 0.4,0.6,0.8,1.0},
    ytick={10, 30, 50},
    legend pos=north west,
    legend style={at={(0.5,-0.55)},
      anchor=north, font=\tiny, legend columns=-1},
    ymajorgrids=true,
    grid style=dashed,
]
\addplot[
    color=green!60!black,
    mark=square,
    ]
    coordinates {
    (0.2, 42.3)  (0.4, 36.4) (0.6, 30.5)  (0.8, 26.1)  (1.0, 24.3)
    };
\end{axis}
\end{tikzpicture}
\end{subfigure}
\begin{subfigure}{0.23\textwidth}
\begin{tikzpicture}
\small
\begin{axis}[
    width=4cm,
    height=3cm,
    xlabel={$t$ },
    ylabel={F1$_{0.8}$(\%)},
    xmin=0.2, xmax=1,
    ymin=5, ymax=70,
    xtick={0.2, 0.4,0.6,0.8,1.0},
    ytick={10, 30, 50, 70},
    legend pos=north west,
    legend style={at={(0.5,-0.55)},
      anchor=north, font=\tiny, legend columns=-1},
    ymajorgrids=true,
    grid style=dashed,
]
\addplot[
    color=green!60!black,
    mark=square,
    ]
    coordinates {
    (0.2, 34.6)  (0.4, 38.9) (0.6, 42.3)  (0.8, 46.9)  (1.0, 46.4)
    };

\end{axis}
\end{tikzpicture}
\end{subfigure}
\hfill
\begin{subfigure}{0.23\textwidth}
\begin{tikzpicture}
\small
\begin{axis}[
    width=4cm,
    height=3cm,
    xlabel={$t$},
    ylabel={Correctness(\%)},
    xmin=0.2, xmax=1,
    ymin=12, ymax=15,
    xtick={0.2, 0.4,0.6,0.8,1.0},
    ytick={12,13,14,15},
    legend pos=north west,
    legend style={at={(0.5,-0.55)},
      anchor=north, font=\tiny, legend columns=-1},
    ymajorgrids=true,
    grid style=dashed,
]
\addplot[
    color=green!60!black,
    mark=square,
    ]
    coordinates {
    (0.2, 13.6)  (0.4, 14.1) (0.6, 13.3)  (0.8, 13.5)  (1.0, 12.9)
    };
\end{axis}
\end{tikzpicture}
\end{subfigure}
\caption{Calibration varies with temperature scaling.}
\label{fig: scaling the temperature}
\end{figure}
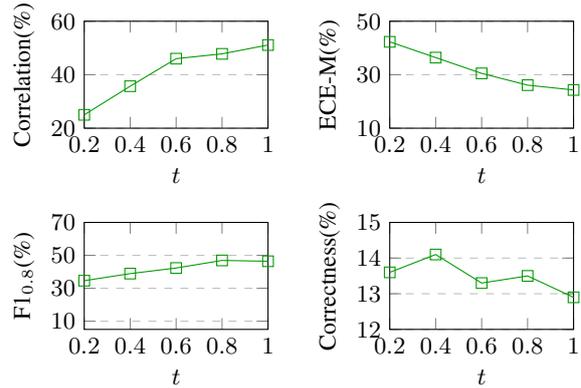

\noindent\textbf{Temperature}
We adjust the generation temperature for Llama2-13b from 0.2 to 1 to examine its impact on calibration. The result in Figure \ref{fig: scaling the temperature} reveals consistent improvements in all calibration metrics. Notably, the model's performance initially improves and then deteriorates.
This observation implies that modulating the generation temperature can enhance the calibration of the model.

\subsection{Application}
\begin{figure}[h]
\includegraphics[width=\columnwidth]{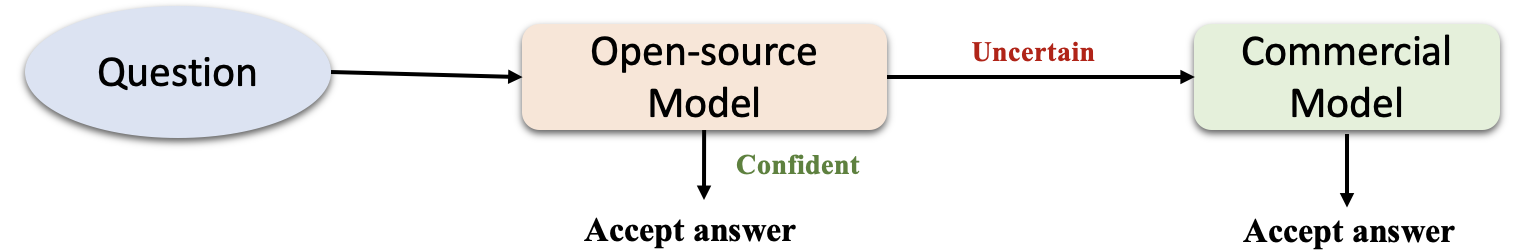}
\caption{The illustration of LLM Cascade.}
\label{fig: llm_cascade}
\end{figure}

\begin{figure}[t]
\centering
\begin{subfigure}{.49\columnwidth}
  \centering
  \small
  \includegraphics[width=1.0\textwidth]{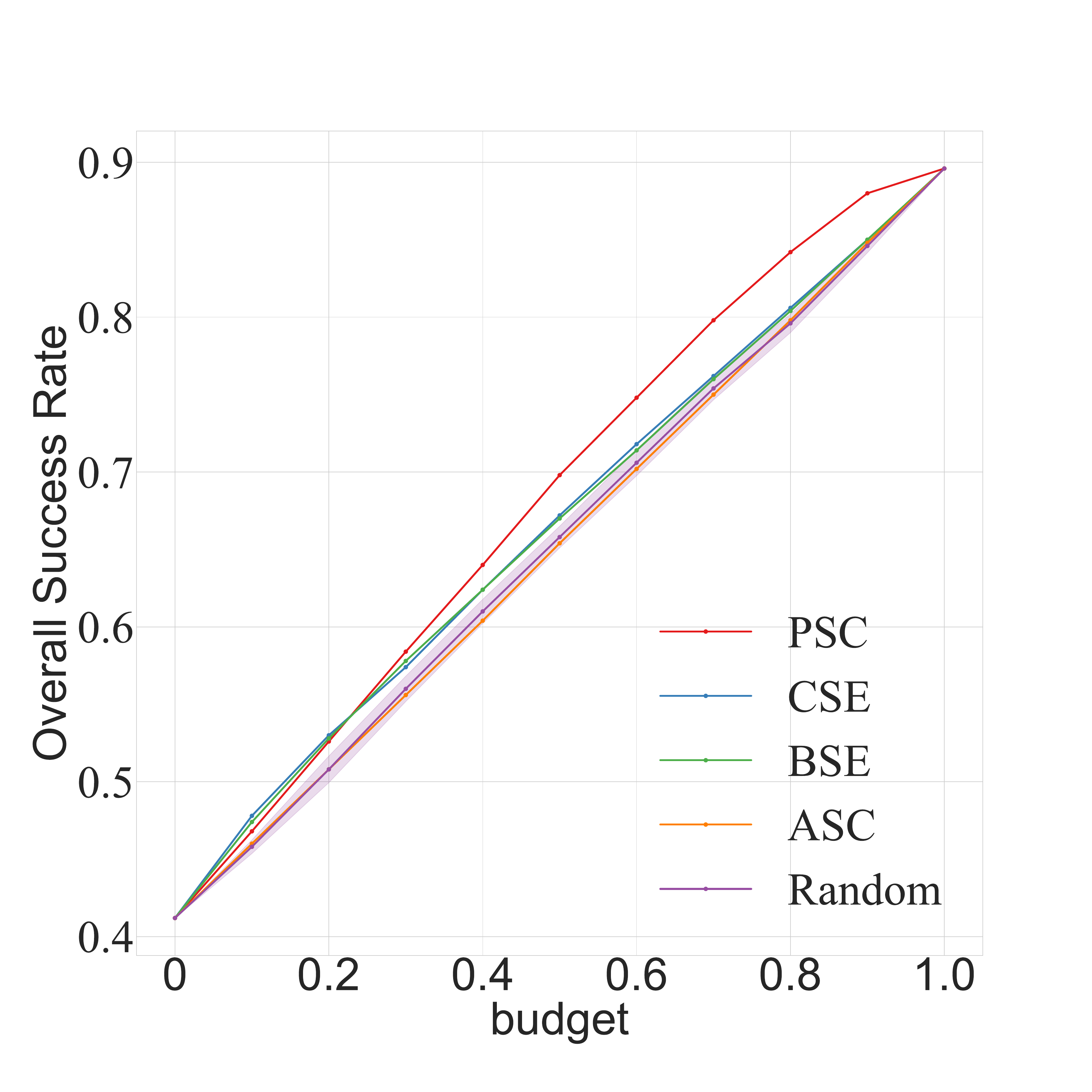}
  \caption{ASQA Overall}
  \label{fig:cascade_asqa_overall}
\end{subfigure}%
\begin{subfigure}{.49\columnwidth}
  \centering
  \small
  \includegraphics[width=1.0\linewidth]{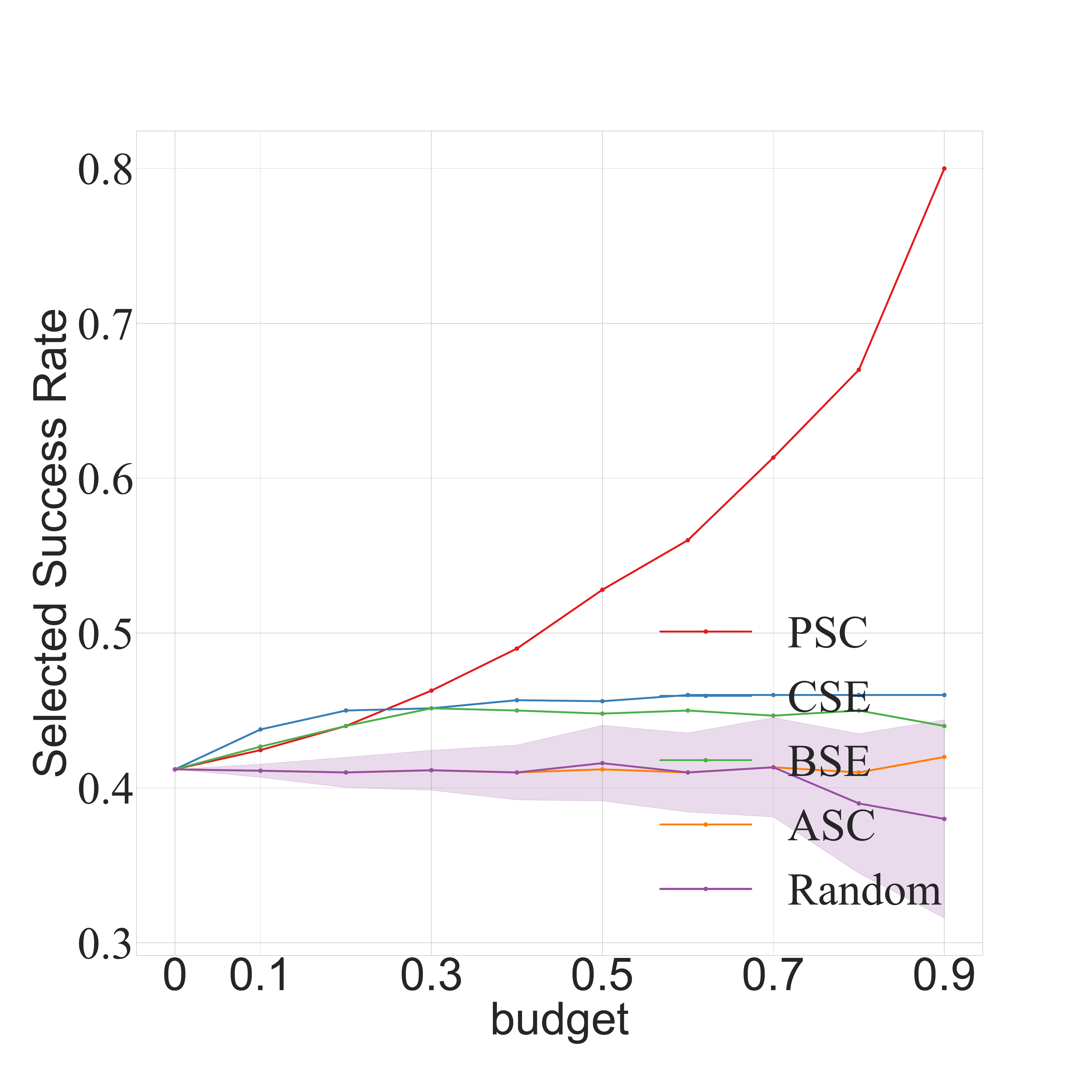}
  \caption{ASQA Selected}
  \label{fig:cascade_asqa_selected}
\end{subfigure}
\\
\begin{subfigure}{.49\columnwidth}
  \centering
  \includegraphics[width=1.0\textwidth]{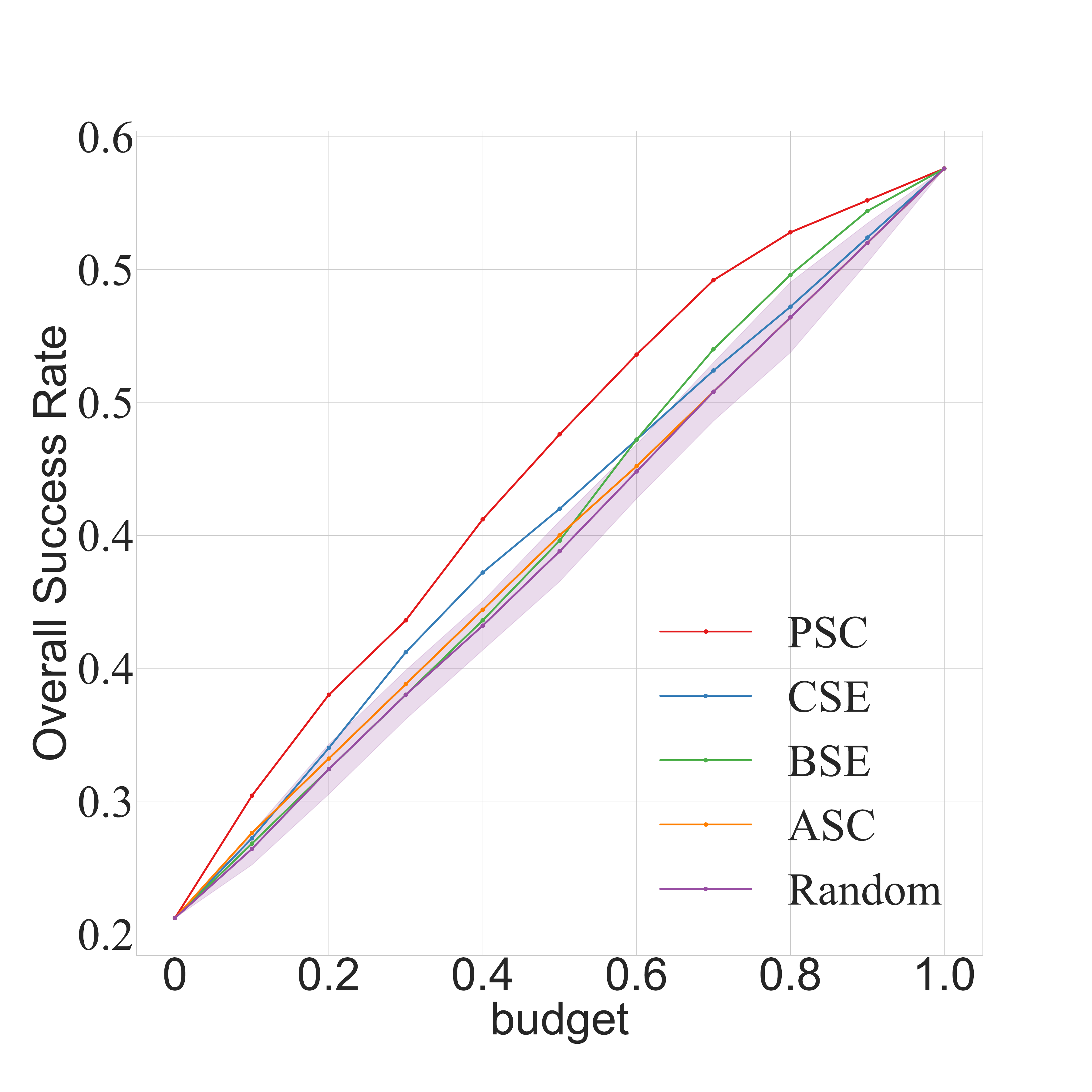}
  \caption{QAMPARI Overall}
  \label{fig:cascade_qampari_overall}
\end{subfigure}
\begin{subfigure}{.49\columnwidth}
  \centering
  \includegraphics[width=1.0\textwidth]{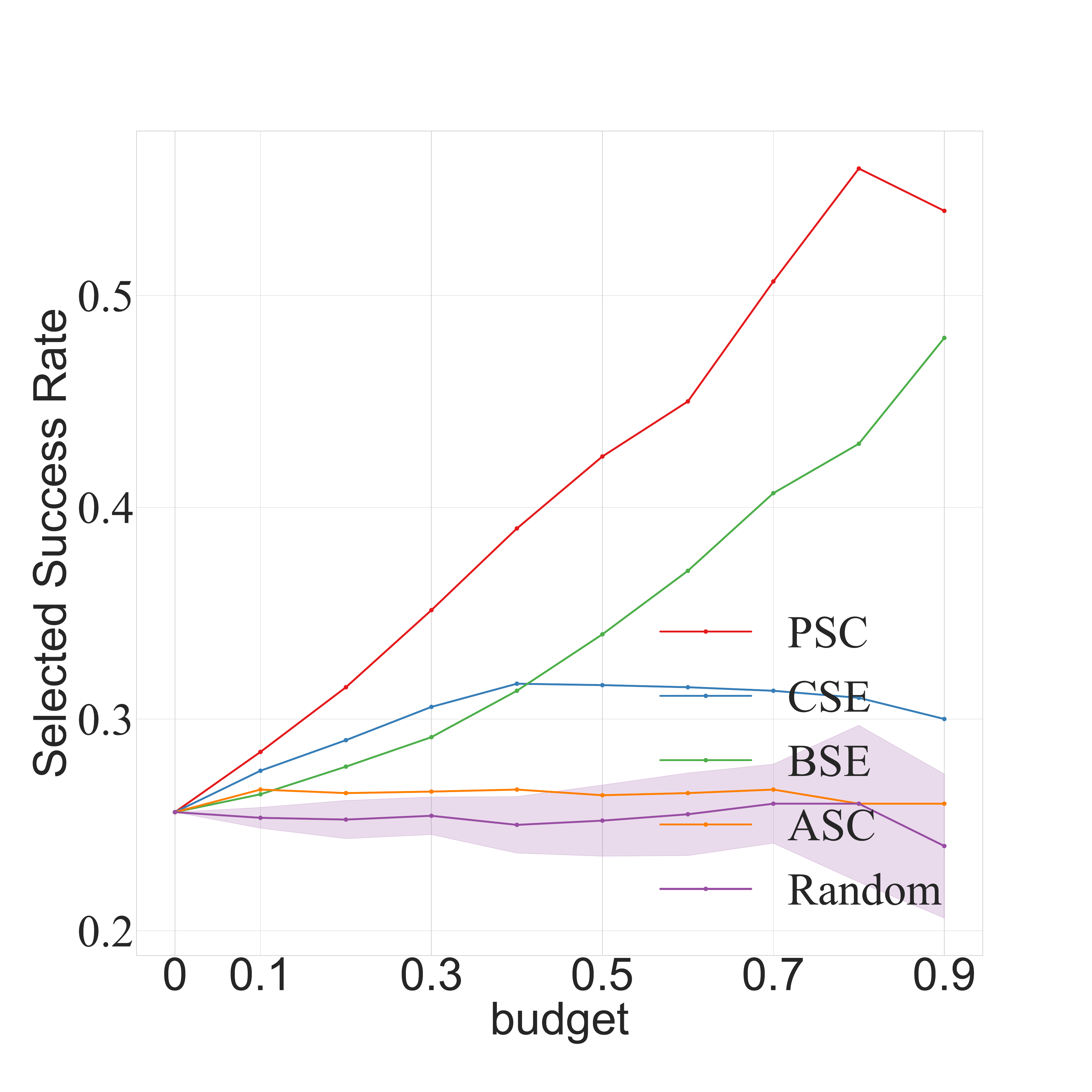}
  \caption{QAMPARI Selected}
  \label{fig:cascade_qampari_selected}
\end{subfigure}
\caption{Variation in Success Rate by API Budget Allocation on the ASQA and QAMPARI Datasets for All Queries and Those Selected by Llama-2.}
\label{fig:llm_cascade_results}
\end{figure}

We showcase an application of long-form calibration in Figure \ref{fig: llm_cascade}: a cost-effective cascading strategy using language models of varying capabilities to efficiently handle queries within an API budget constraint.
Initially, an open-source model (Llama-2-13b in our experiment) address questions where it believes the answer has a probability higher than $\tau_c$ that the answer's correctness score is above $\tau_s$. Complex queries, flagged by lower model confidence, are escalated to a more advanced API LM (GPT-4). Adjusting $\tau_c$ between 0 and 1 controls how many queries reach GPT-4, balancing answer quality with API budget constraints. We benchmark using the open-source LM for a zero API budget and the commercial LM for full-budget scenarios. 
Our experiments utilize the ASQA and QAMPARI datasets to evaluate four distinct confidence elicitation strategies: PSC, ASC, CSE, and BSE. Additionally, we incorporate a baseline strategy where, under a constrained number of API requests, a random selection of queries is processed by Llama-2-13b, with the remaining handled by GPT-4. For each API budget scenario, questions are randomly assigned to Llama-2-13b using 10 different random seeds, and we calculate the mean and standard deviation of the results. We focus on the success rate, which we define as the percentage of answers that meet or exceed a user-specified score threshold. This metric is reported both for the overall dataset and for the subset of queries selected and handled by Llama-2-13b, illustrating both the general effectiveness of our cascading model and the selective answering capabilities of the individual model. As shown in Figure \ref{fig:llm_cascade_results}, PSC generally outperforms the other methods, with CSE and BSE yielding comparable results that  follow. ASC, in contrast, performs the poorest, comparable to the random selection strategy. These results highlight the pivotal role of advanced calibration techniques, confidence elicitation methods in our case, in boosting the practical utility and cost-efficiency of LLMs when API usage is limited.

\section{Conclusion}
Our study presents a novel calibration system for evaluating LLMs in long-form generation. Our results challenge the assumption that larger LLMs are always calibrated better and show calibration variability across datasets. Additionally, we propose methods to improve LLM calibration and show an application that optimizes performance under API budget constraints.  The system we present is crucial for further improving the liability of LLMs.

\section{Limitation}
Our study faces three primary limitations. First, we rely on GPT-4 to estimate the target correctness distribution. But as tasks become more subjective, consensus on humans' evaluations may decrease.  This wider target correctness distribution becomes challenging for both GPT-4 and human annotators to accurately capture. This limitation is inherent to natural language generation (NLG) evaluation and lies beyond the purview of our project. Our framework operates under the premise that a target correctness distribution exists and concentrates on calibration which aligns the model's confidence with this assumed target. Second, our experiments focus on long-form QA and do not extend to specialized domains such as law, medicine, or education, where the calibration of LLMs could have significant real-world implications. Lastly, our self-consistency method is computationally intensive, posing a challenge for practical applications. There is a need for more efficient approaches in real-world settings.
\\ \\
\textbf{Replicability}:\\
Codes: \url{https://github.com/kkkevinkkkkk/calibration}

\bibliography{anthology,custom}

\appendix
\section{Dataset}
\label{appendix:dataset}
\textbf{ASQA} (Answer Summaries for Questions which are Ambiguous) \citep{stelmakh-etal-2022-asqa} is a specialized long-form factoid dataset, designed to address ambiguous factoid questions that yield different correct answers based on their interpretations. This dataset challenges models to synthesize factual information from multiple sources, creating coherent long-form summaries that effectively resolve the inherent ambiguities in these questions.
\\ \\
\textbf{ELI5} \citep{fan-etal-2019-eli5} is a comprehensive open-ended long-form dataset, encompassing over 270,000 threads from the Reddit forum ``Explain Like I'm Five.'' This unique platform features community-generated responses to a wide array of questions, all tailored to be easily understandable by a five-year-old audience. The majority of queries in ELI5 are centered around 'how,' 'why,' and 'what' questions, which necessitate comprehensive, detailed responses supported by evidence from multiple passages.
\\ \\
\textbf{QAMPARI} \citep{Amouyal2022QAMPARIAO} is a factoid dataset where answers are presented as lists of entities dispersed across multiple paragraphs. Its construction involves an automated process that utilizes Wikipedia knowledge graphs and tables. Questions are manually paraphrased, and answers are thoroughly verified for accuracy. Notably, each question in QAMPARI is associated with an average of 13 answers, demonstrating its breadth.
\\ \\
\textbf{CNNDM}~\citep{nallapati-etal-2016-abstractive} is a large-scale news summarization dataset containing news articles from CNN\footnote{\url{https://www.cnn.com/}} and DailyMail\footnote{\url{https://www.dailymail.co.uk/}}.
The original CNNDM dataset consists of both source news articles and reference summaries.
However, recent work~\cite{liu-etal-2023-revisiting, Zhang2023BenchmarkingLL} has found that the provided reference summaries are not of very good quality and zero-shot LLMs summaries are preferred by human annotators over the reference summaries.
\\ \\
In the \textbf{ACLE} (Automatic LLMs' Citation Evaluation) \citep{gao-etal-2023-enabling}, a pioneering benchmark for assessing LLMs' citation capabilities, a subset of 1,000 examples is randomly selected from the development sets of ASQA, ELI5, and QAMPARI to form a test set for each task. For our specific analysis, we choose to utilize the first 500 examples from each of these datasets in ACLE as our test set, providing a focused and representative sample for each task. For CNNDM, we utilize 100 examples as our test set.

\section{Self-consistency}
\label{appendix: self-consistency}
We propose four different self-consistency based methods tailored for long-form generation, each with a different strategy to measure the similarity between two long-form answer.
\\ \\
\textbf{Naive}
The most basic approach utilizes an additional LLM (GPT-3.5-turbo in our experiments, which can be replaced by other models trained for this task) to determine if two responses are akin, assigning a corresponding similarity score. This method diverges from the relevant technique in contemporary research \citep{Chen2023UniversalSF}, which primarily focuses on identifying the most consistent answer. Instead, our approach aims to secure specific consistency ratings that reflect the model's assurance in its primary answer, offering a general overview of the answers' similarity. The template for similarity measuring can be found in Appendix \ref{subsec: self-consistency template}.
\\ \\
\textbf{Sentence Split}
For a more detailed similarity analysis between the two answers, we split the first answer into individual sentences. Another LLM (GPT-3.5-turbo or a similar NLI model) is then used to evaluate whether similar statements are present in the second answer. This method's limitation is that not all sentences carry equal informational weight. Some may be filler or less informative, potentially skewing the similarity assessment. The template can be found in Appendix \ref{subsec: self-consistency template}.
\\ \\
\textbf{Claim}
To further refine the approach, we focus on sentences that make factual claims. This involves two steps: first, using a claim detector to identify factual claims within a sentence and then using an NLI model to determine if similar factual claims exist in the second answer. This method operates under the assumption that factual claims are the most critical components of an answer, representing its core information. We leverage a DeBERTa-V2 \citep{He2020DeBERTaDB} trained by ClaimBuster as our fact detector and GPT-3.5 as the NLI models. 
\\ \\
\textbf{Named Entity Recognition}
Advancing the granularity further, we compare named entities between two responses. We identify and compare entities present in both answers by utilizing a named entity recognition model. The degree of overlap in these entities serves as an indicator of answer similarity. This approach focuses more on concrete, identifiable elements within the answers. We use a Roberta-large \citep{Liu2019RoBERTaAR} trained with SpanMarker framework \footnote{\url{https://github.com/tomaarsen/SpanMarkerNER}}, which can be replaced by other NER models.

\section{Evaluation Metric}
\label{appendix: evaluation metric}
\subsection{GPT-4 metric}
We ask GPT-4 to range an answer from 0 (worst) to 5 (best), which is then normalized to [0, 1]. See Appendix \ref{appendix: correctness evaluation template} for details of the template. 
To mitigate scoring variability from criteria ambiguity and LLM uncertainty, we have evaluators repeatedly score each answer, forming a score distribution that better reflects its correctness. Specifically, given an LLM evaluator $\mathcal{E}$, $N$ evaluations from it, and a score $x\in [0, 1]$, the correctness distribution is:
\begin{align}
    P_{T_i}(x) &= \frac{1}{N}\sum_{j=1}^N \mathrm{1}(\mathcal{E}(A_i)_j = x)
\end{align} 

\subsection{Human Evaluation}
\label{appendix: human evaluation}
We utilize GPT-4 to assess answers across different tasks, including ASQA, ELI5, and CNNDM, using GPT-4 scores as a proxy for the target distribution. To demonstrate the better alignment of GPT-4 scores with human preferences over task-specific metrics, we focus our human evaluation efforts on the long-form tasks of ASQA and ELI5. This approach is supported by prior research indicating GPT-4's congruence with human judgments on summarization tasks \citep{liu-etal-2023-g}, thereby obviating the need for manual evaluation of CNNDM. Following \citet{gao-etal-2023-enabling}, in ASQA, we adopt the \textit{EM-recall} automatic metric, which gauges the recall of correct short answers by verifying if the dataset's provided short answers are exact substrings of the generated content, following established methodologies. For ELI5, we utilize the most precise automatic metric to date, \textit{claim recall}, employing the TRUE \citep{honovich-etal-2022-true-evaluating} natural language inference model to ascertain if the generated output encompasses the sub-claims of the reference answer.

\begin{figure*}[htbp]
\centering
\begin{minipage}{0.3\textwidth}
  \includegraphics[width=\linewidth]{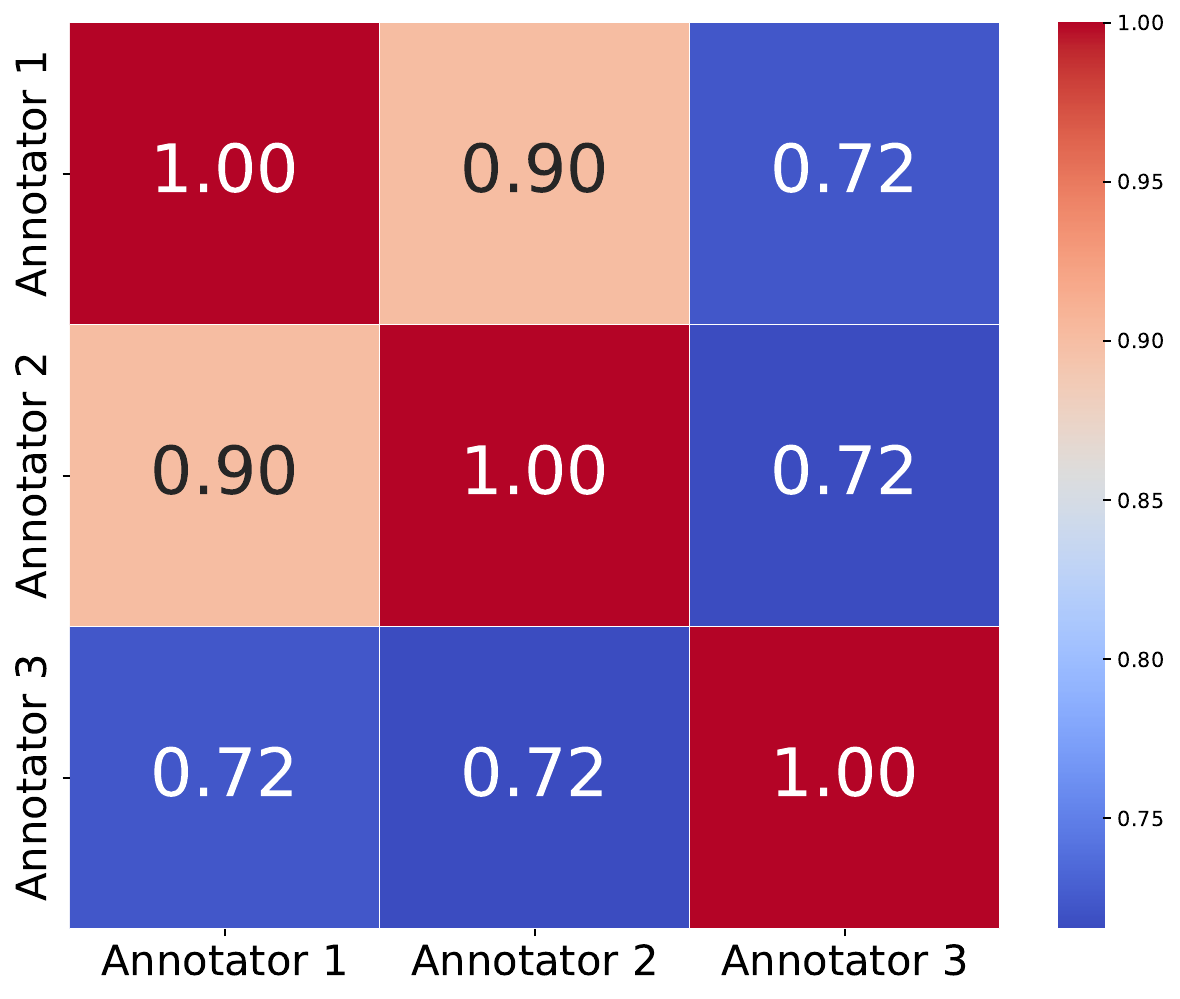}
  \caption{Correlation for ASQA}
\end{minipage}\hfill
\begin{minipage}{0.3\textwidth}
  \includegraphics[width=\linewidth]{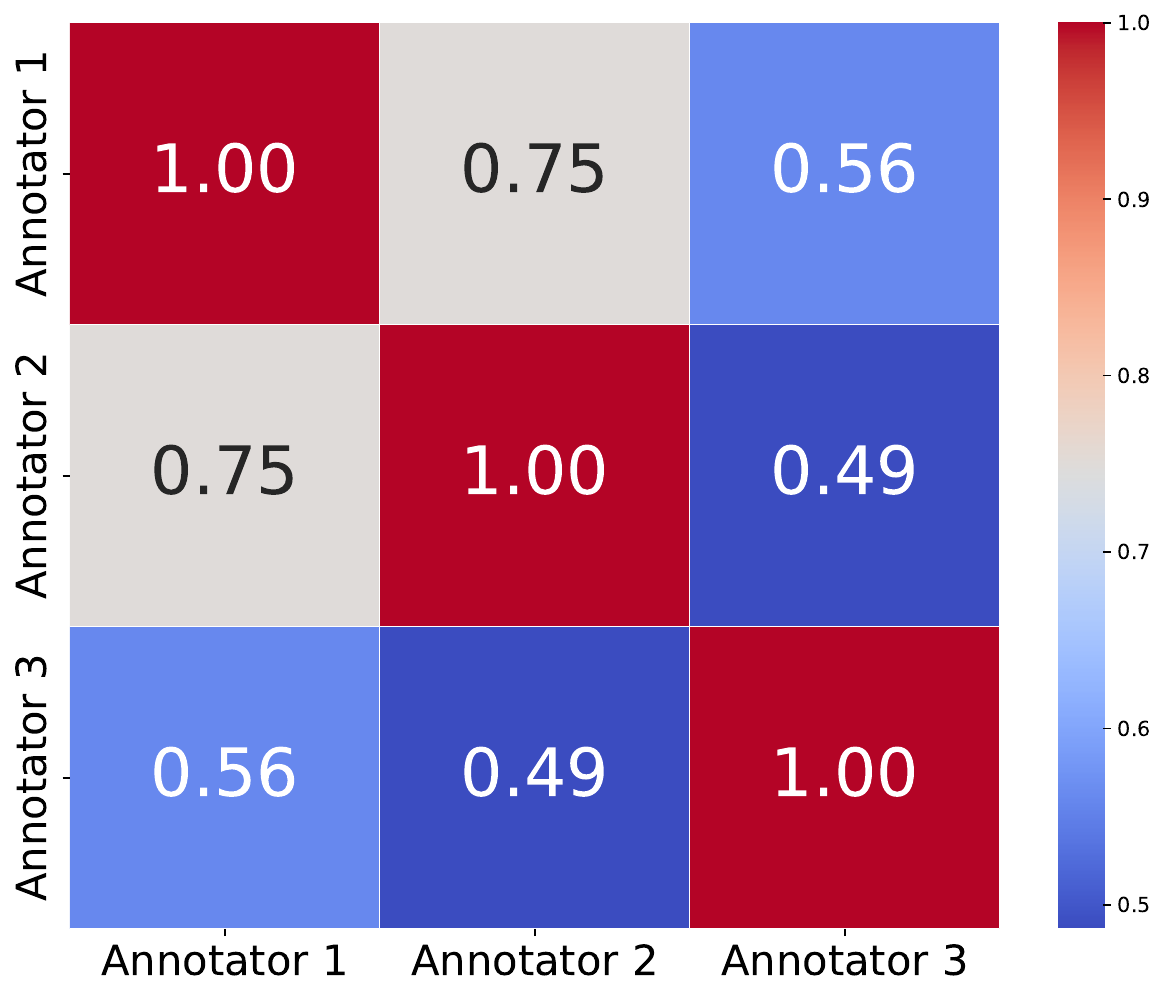}
  \caption{Cohen Kappa for ASQA}
\end{minipage}\hfill
\begin{minipage}{0.3\textwidth}
  \includegraphics[width=\linewidth]{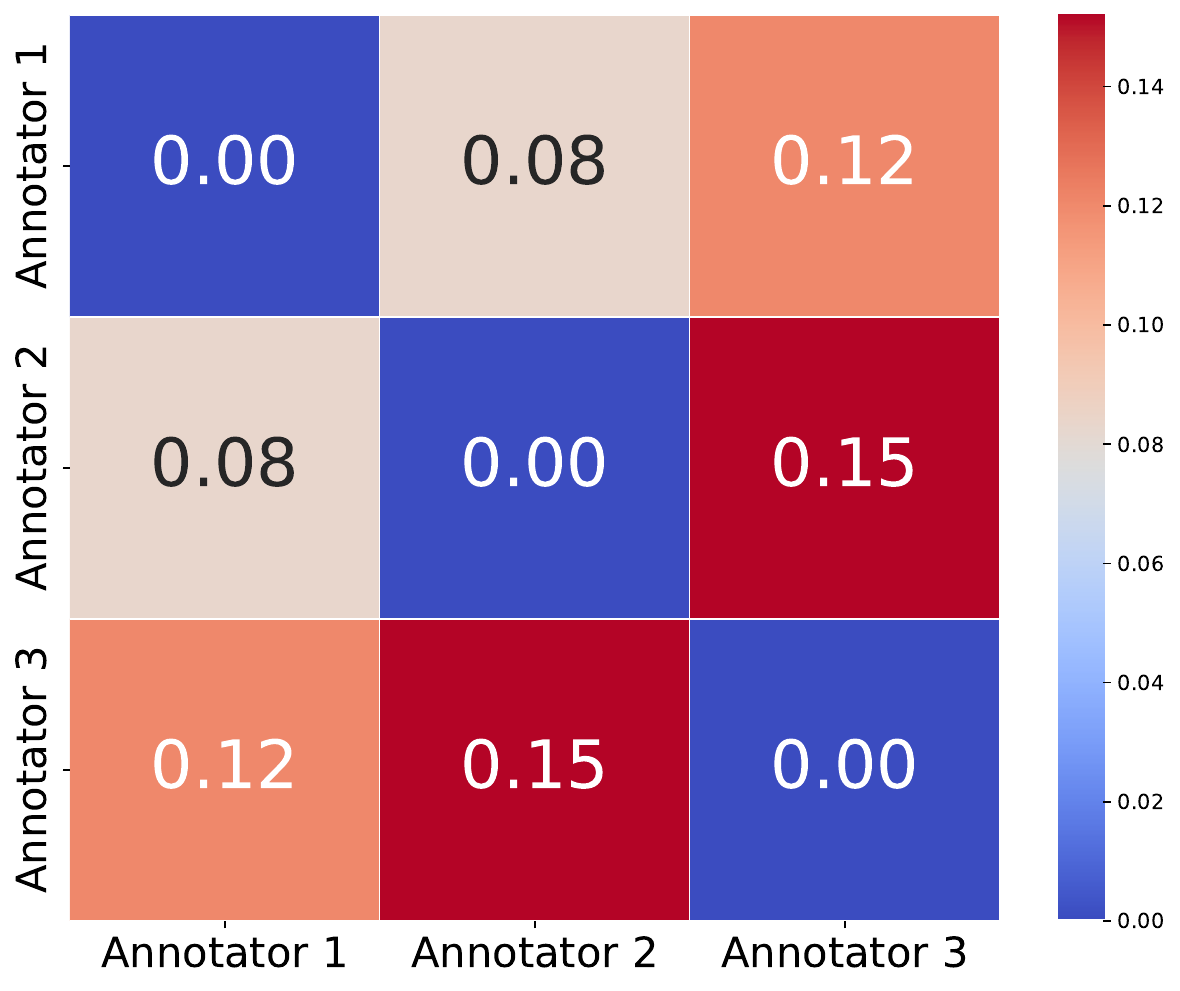}
  \caption{MAE for ASQA}
\end{minipage}

\vspace{1cm} %

\begin{minipage}{0.3\textwidth}
  \includegraphics[width=\linewidth]{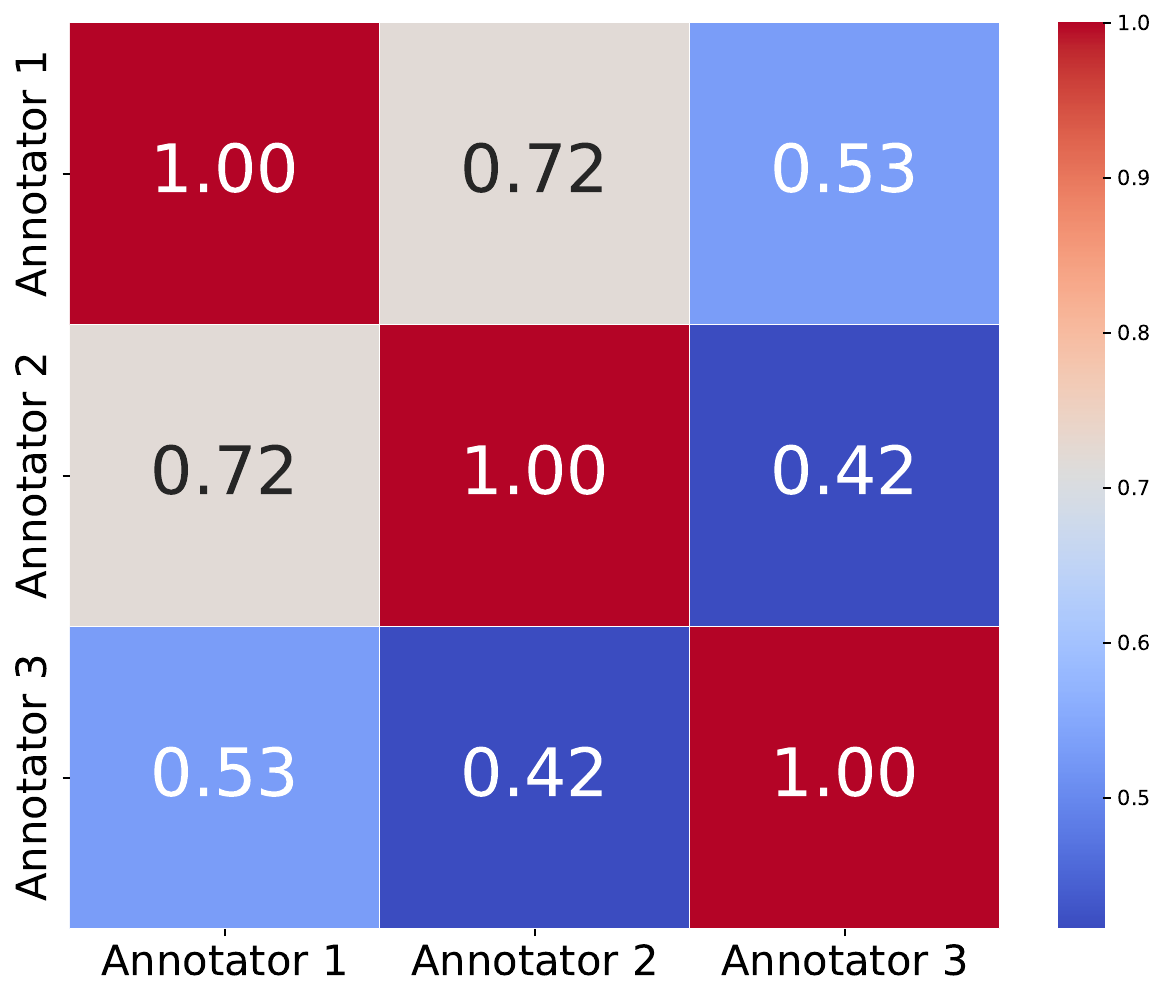}
  \caption{Correlation for ELI5}
\end{minipage}\hfill
\begin{minipage}{0.3\textwidth}
  \includegraphics[width=\linewidth]{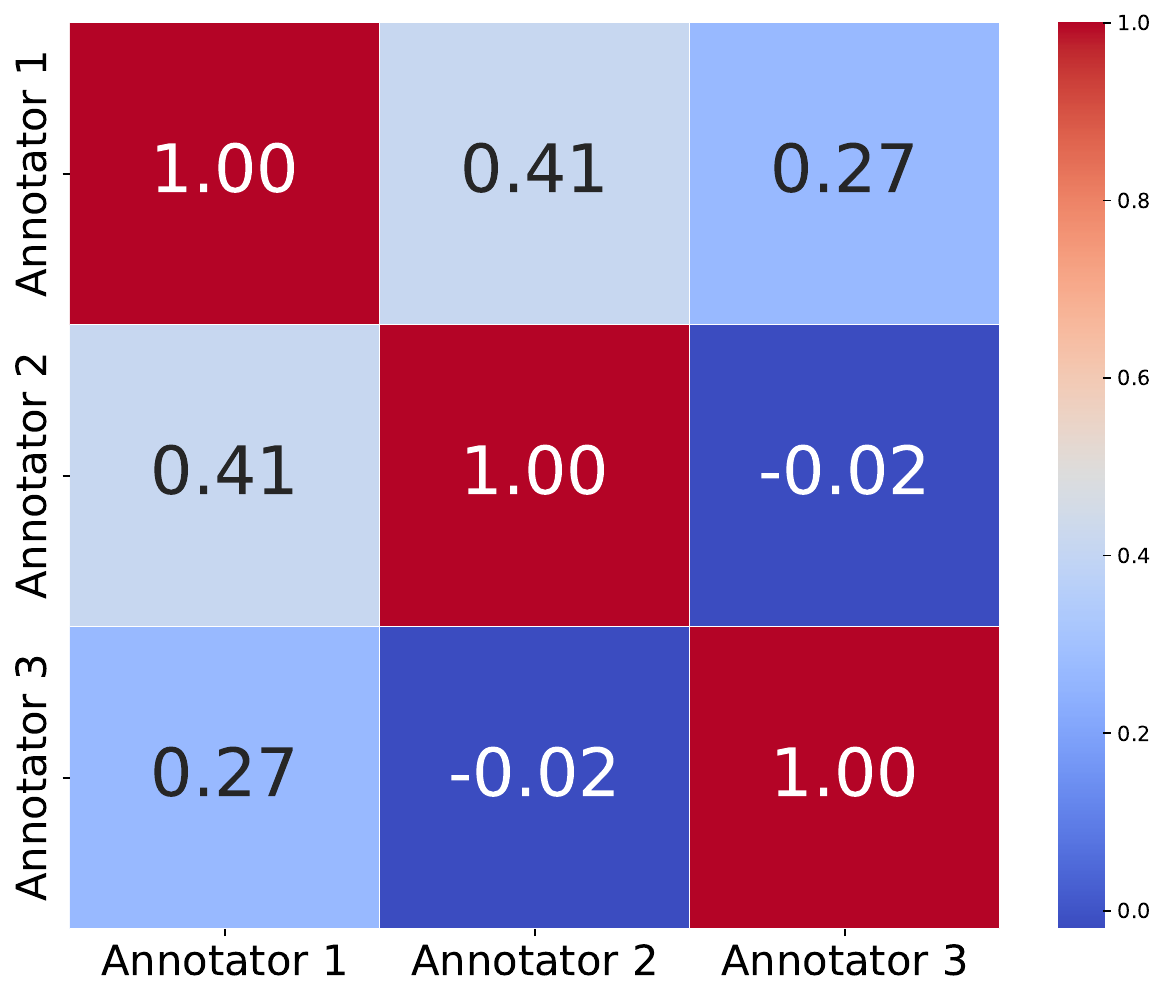}
  \caption{Cohen Kappa for ELI5}
\end{minipage}\hfill
\begin{minipage}{0.3\textwidth}
  \includegraphics[width=\linewidth]{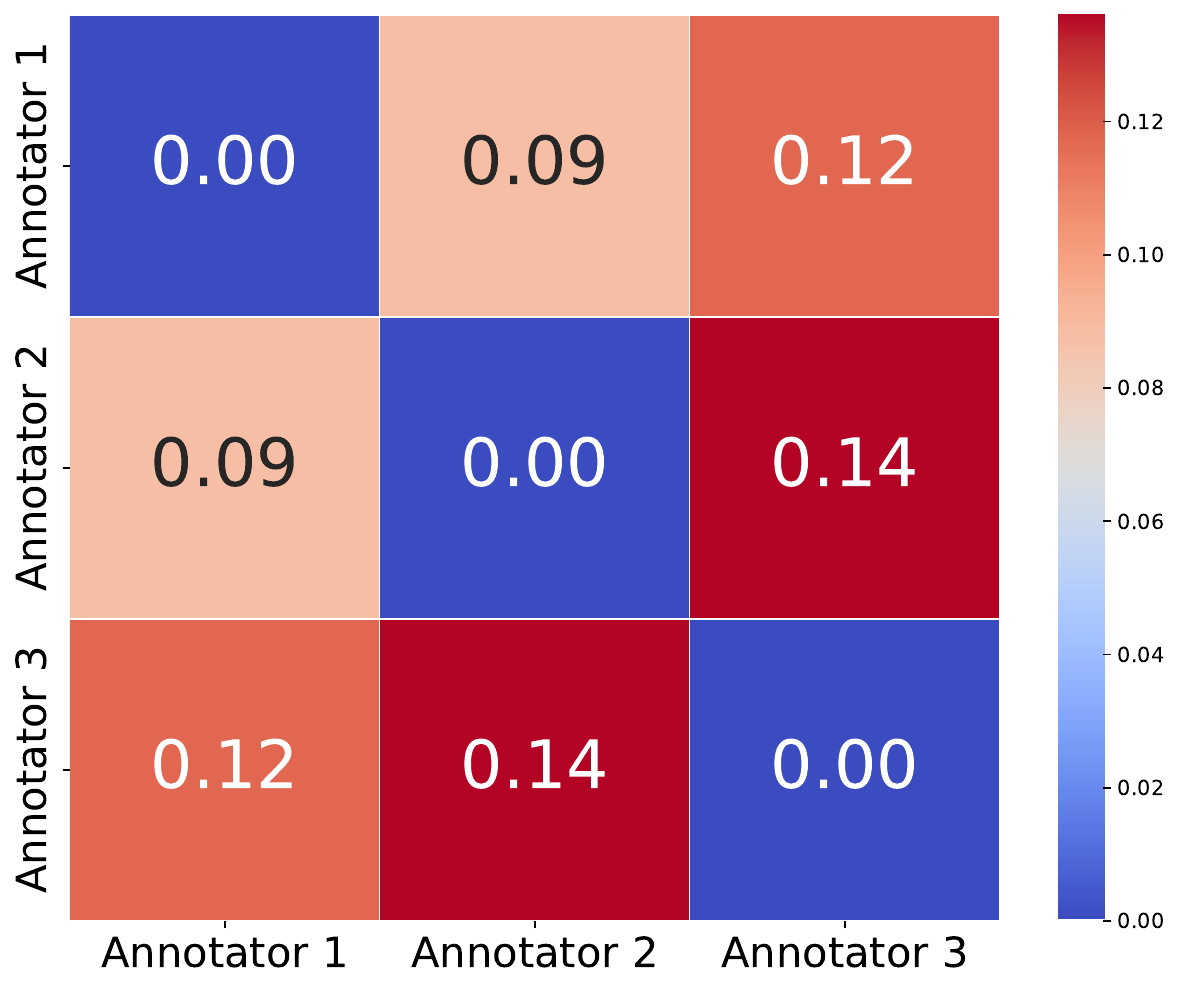}
  \caption{MAE for ELI5}
\end{minipage}
\caption{Pairwise annotator agreements for ASQA and ELI5 respectively}
\label{fig: annotator agreement}
\end{figure*}

We present the task criteria to humans and ask them to provide a score for each answer based on the criteria. Participants are provided with a reference answer—not as an exclusive ground truth but as a guide—and are permitted to use search engines for additional context. We enlisted three annotators to evaluate 75 samples each for ASQA and ELI5, and calculate the average of them as a human score. 
Our analysis compares these human scores with those generated by task-specific metrics and GPT-4. The results, as detailed in our Table \ref{tab: human eval}, underscore GPT-4's closer alignment with human judgments in both ASQA and ELI5.
As shown in Figure \ref{fig: annotator agreement}, as the task becomes more open-ended like Eli5, the human agreements become lower than the factoid dataset ASQA. This further evidence the assumption that the correctness of a long-form answer should be a distribution.
\begin{table}[htb!]
\centering
\small
\begin{tabular}{l|cc|cc}
\toprule
&  \multicolumn{2}{c|}{\cellcolor{gray!15}\textbf{ASQA}} & \multicolumn{2}{c}{\cellcolor{gray!15}\textbf{ELI5}} \\
  \textbf{Metric} & \textbf{EM} & \textbf{GPT-4} & \textbf{Claim} & \textbf{GPT-4}\\ 
\midrule
Corr $\uparrow$ & 47.8 & 76.2 & 42.9 & 71.5 \\ 
MAE $\downarrow$ & 43.0 & 12.9 & 52.0 & 9.1 \\
\bottomrule
\end{tabular}
\caption{Comparison of Human, Task-Specific metric, and GPT-4 Correctness Distributions in ASQA and ELI5 Tasks. Results are with \%.  `Corr' denotes Pearson correlation (the higher the better), `MAE' denotes mean absolute error (the lower the better). `EM' denotes EM-recall, and `Claim' denotes Claim-recall.}
\label{tab: human eval}
\end{table}

\section{Main Experiment}
\label{appendix:experiment_details}
We assess the calibration of variously sized models (Llama-2-13b, Llama-2-70b, Vicuna-13b, GPT-3.5-turbo) across three long-form QA datasets (ASQA, ELI5, QAMPARI) and one summarization dataset (CNNDM). The process involves several steps: 1. Generation: Asking the model to generate answers for questions in the dataset 2. Correctness Assessment: We utilize GPT-4 to evaluate the correctness of the models' answers, except for QAMPARI, where we directly apply the F1-5 metric. 3. Confidence Distribution Derivation: After generating answers, models employ self-evaluation or self-consistency methods to derive their confidence distribution. 4. Calibration Measurement: .
\subsection{Generation}
In the answer generation phase, we employ a 3-shot in-context learning approach for each long-form QA dataset, providing three exemplars to guide the models. For the CNNDM dataset, we adopt a 0-shot strategy, aligning with prior studies \citep{Goyal2022NewsSA}. The generation of answers utilizes a top-K sampling method, setting the generation temperature for all models at 0.6 and the top-K parameter at 10.

\subsection{Correctness Evaluation}
    For ASQA, ELI5, CNNDM, we ask the GPT-4 to evaluate the correctness of examples. For ASQA and ELI5, We ask GPT-4 to evaluate the answer three times, producing a distribution capturing criteria ambiguity and model's inherent subjectivity. However, for CNNDM, we only ask GPT-4 to evaluate once to save computation given that most answers correctness concentrates around 0.8 with small variance.
    The GPT-4 evaluation template can be found in Appendix \ref{appendix: correctness evaluation template} and criteria for different tasks can be found in Appendix \ref{appendix: criteria}. For QAMPARI, which involves generating a list of entities as answers, we determine that F1 scores provide a more suitable measure of correctness than GPT-4 evaluations. To this end, we adopt the F1-5 metric \citep{gao-etal-2023-enabling}, computing the F1 score based on direct matches with the correct answer list and assigning a recall of 100\% for responses containing at least five accurate answers (recall-5).
\subsection{Confidence Elicitation}
\label{Appendix: confidence elicitation}
After getting the answer from the model, we leverage the self-evaluation or self-consistency method to derive confidence distribution from the model. 
\\ \\
\textbf{Self-Evaluation}
We prompt the model to self-evaluate an answer 10 times, creating a confidence distribution based on its self-evaluation scores. Each self-evaluation includes task instructions, grading criteria, and evaluation examples (three for all long-form QA datasets and one for CNNDM due to length constraints). Additionally, it contains specific instructions for self-evaluation, incorporating both the question and the answer under evaluation. The detailed self-evaluation template and criteria are available in Appendix \ref{subappendix: self-evaluation template} and Appendix \ref{appendix: criteria} respectively.
\\ \\
\textbf{Self-consistency}
In the self-consistency approach, we generate an answer to the same question 10 times, designating the first response as the primary answer. We then calculate similarity scores between the primary answer and the remaining responses. These scores collectively create the model's confidence distribution for the primary answer. For various datasets, tailored strategies are employed to compare similarities between two answers.

For ASQA, we employ a self-consistency-claim approach. To assess the similarity between the two answers, we first identify factual claims in the first answer using a ClaimBuster-trained DeBERTa V2 \citep{He2020DeBERTaDB} claim detector. We then verify the presence of these claims in the second answer through NLI. The similarity score is the average presence of factual claims across answers.

In QAMPARI, self-consistency-NER is used to determine confidence distribution. As the answers are entity lists, we extract entities by separating commas. The similarity score is calculated based on the proportion of overlapping entities between two answers, relative to the total entities in the first answer.

For ELI5, where answers provide easily understandable explanations, we focus on the overall content. Here, we apply a self-consistency-naive method, assign a similarity score to each answer pair with simple prompting. 

In CNNDM, where answers are summaries highlighting key points of an article, we gauge the similarity between two answers by evaluating the overlap of key points. To achieve this, we implement the self-consistency-split method. We dissect the first answer into individual sentences and then use NLI to determine if each sentence is present in the second answer. The similarity score is derived by averaging the presence of these segmented sentences in the comparative answer.

\section{Improving Calibration}
\subsection{Fine-tuning}
\label{appendix:fine-tuning}
In our study, we focus on improving model calibration on the ASQA dataset. We explore three different strategies: fine-tune the model to do self-evaluation (Input: question and model's answer; Output: answer's score and explanation), fine-tune the model to do generation (Input: question; Output: answer), and a hybrid of both.
\\ \\
\textbf{Data}
We generate a self-evaluation training dataset with GPT-4, which evaluates different models responses to questions drawn from ASQA training set. This dataset comprises inputs of questions and corresponding model answers, with outputs including a score and an explanation for each answer. We create the self-evaluation data in two steps. The initial phase involves the creation of a diverse answer pool. This is achieved by employing a suite of models with varying computational capacities, including Llama-2-7b, Llama-2-13b, Llama2-70b, Vicuna-13b, and ChatGPT. Each model generates responses to a spectrum of questions drawn from the training set of ASQA task, ensuring the resultant answer pool encompasses a broad quality spectrum, from low to high. Subsequently, we employ GPT-4 to critically assess these answers, assigning a score and providing a corresponding explanation for its evaluation. This process yields a rich dataset, each instance of which encompasses a question, a model-generated answer, an evaluative score, and a justification for that scoring. This approach results in a comprehensive dataset with 2,000 self-evaluation examples (1,800 for training and 200 for validation), each including a question, model-generated answer, score, and justification. 
For generation data, we use 80\% of training ASQA's dataset (4,353 examples) for training and the remaining 20\% for validation. The hybrid dataset combines the self-evaluation and generation training sets, using the self-evaluation validation set for assessment.
\\ \\
\textbf{Training and Results}
Regarding training, we fine-tune Llama2-13b-chat model using LORA on this dataset. We maintain consistent parameters across all scenarios: a learning rate of 5e-6, five epochs, 100 warm-up steps, and a total batch size of 4 (achieved through 4 gradient accumulation steps across four GPUs, with a batch size of 1 per device). 
As Table \ref{tab:fine-tuning} reveals, solely training on self-evaluation (`Evaluation') did not yield consistent improvements in calibration, possibly due to the complexity of this task, as well as the limitation of LORA fine-tuning. Nonetheless, fine-tuning the model improves the self-consistency method, especially when the generation data is included during training (`Eval + Gen' and `Generation'). The model becomes more confident in terms of self-consistency after fine-tuning.

\subsection{Source Documents}
\label{appendix: source documents}
We investigate the effect of additional context on LLM calibration, focusing on self-consistency confidence with Llama-2-13b-chat and GPT-3.5-turbo models on the ASQA dataset. We test the model's performance when supplemented with two different types of source documents: random documents related to the question and `oracle' documents directly relevant to the answers, as included in the dataset release \cite{stelmakh-etal-2022-asqa}. 
Findings in Table \ref{tab:source_documents} reveal that Oracle documents can enhance model performance and calibration across two out of three metrics for both models, while random documents are less effective.
These results underscore the importance of relevant contextual information in model calibration.
\begin{table}[h!]
\small
\centering
\begin{tabular}{lccccc}
\toprule
\textbf{Model} &\textbf{Doc} & \textbf{Corr} & \textbf{ECE-M} & \textbf{F1$_{0.8}$} & \textbf{Score} \\ 
\midrule
\multirow{3}{*}{Llama2$_{13b}$} & N & 48.0\textcolor{white}{$-$} & 15.9\samewhite & 3.6\textcolor{white}{$-$} & 51.3\textcolor{white}{$-$}\\ 
& R & 39.0 \textcolor{red!60!black}{$\downarrow$} & 14.8\gdown & 30.6 \textcolor{green!60!black}{$\uparrow$} & 59.6 \textcolor{green!60!black}{$\uparrow$}\\ 
& O & 49.5 \textcolor{green!60!black}{$\uparrow$} & 17.5\rup & 20.1 \textcolor{green!60!black}{$\uparrow$} & 65.8 \textcolor{green!60!black}{$\uparrow$}\\ 
\midrule
\multirow{3}{*}{GPT-3.5$_{turbo}$} & N & 30.5\textcolor{white}{$-$} & 26.7\samewhite & 58.4\textcolor{white}{$-$} & 72.6\textcolor{white}{$\downarrow$}\\ 
& R & 36.6 \textcolor{green!60!black}{$\uparrow$} & 26.4\gdown & 68.2 \textcolor{green!60!black}{$\uparrow$} & 64.8 \down\\  
& O & 27.0 \down & 24.9\gdown & 72.8 \textcolor{green!60!black}{$\uparrow$} &  75.8 \textcolor{green!60!black}{$\uparrow$}\\  
\bottomrule
\end{tabular}
\caption{How source documents affect the calibration score. In the document column, ``N'' means no documents in the input prompt, ``R'' means randomly selected documents relevant to the topic, and ``O'' means the oracle documents relevant to the answer. \textcolor{green!60!black}{$\uparrow$} denotes that the calibration score goes up when adding documents, while \textcolor{red!60!black}{$\downarrow$} means going down. For ECE-M it's opposite. }
\label{tab:source_documents}
\end{table}

\subsection{Hybrid Confidence Elicitation}
\label{appendix: hybrid confidence elicitation}
\begin{figure}[htb]
\centering
\hspace*{-0.4cm}%
\vspace{-0.3cm}
\begin{tikzpicture}
\small
\begin{axis}[
    title={Calibration Scores vs. Weight Variation},
    width=8cm,
    height=3cm,
    xlabel={$\alpha$ },
    ylabel={Correlation(\%)},
    xmin=0, xmax=1,
    ymin=40, ymax=60,
    xtick={0, 0.2, 0.4,0.6,0.8,1.0},
    ytick={40, 45, 50, 55, 60},
    legend pos=north west,
    legend style={at={(0.5,-0.55)},
      anchor=north, font=\tiny, legend columns=-1},
    ymajorgrids=true,
    grid style=dashed,
]
\addplot[
    color=green!60!black,
    mark=square,
    ]
    coordinates {
    (0, 56.1) (0.1, 57.1) (0.2, 58.0) (0.3, 58.8) (0.4, 59.2)
    (0.5, 58.4) (0.6, 56.4) (0.7, 52.4) (0.8, 46.3) (0.9, 38.4) (1.0, 29.6)
    };
\addplot[
    color=blue,
    mark=square,
    ]
    coordinates {
    (0, 45.4) (0.1, 46.1) (0.2, 46.7) (0.3, 47.2) (0.4, 47.3)
    (0.5, 46.8) (0.6, 45.0) (0.7, 41.6) (0.8, 35.9) (0.9, 28.4) (1.0, 20.2)
    };
\addplot[
    color=orange,
    mark=square,
    ]
    coordinates {
    (0, 43.9) (0.1, 44.2) (0.2, 44.5) (0.3, 44.7) (0.4, 44.6)
    (0.5, 43.9) (0.6, 42.2) (0.7, 38.9) (0.8, 33.2) (0.9, 25.1) (1.0, 15.8)
    };
\end{axis}
\end{tikzpicture}

\hspace*{-0.4cm}%
\begin{tikzpicture}
\small
\begin{axis}[
    width=8cm,
    height=3cm,
    xlabel={$\alpha$ },
    ylabel={F1$_{0.8}$(\%)},
    xmin=0, xmax=1,
    ymin=5, ymax=70,
    xtick={0, 0.2, 0.4,0.6,0.8,1.0},
    ytick={10, 30, 50, 70},
    legend pos=north west,
    legend style={at={(0.5,-0.55)},
      anchor=north, font=\tiny, legend columns=-1},
    ymajorgrids=true,
    grid style=dashed,
]
\addplot[
    color=green!60!black,
    mark=square,
    ]
    coordinates {
    (0, 35.8) (0.1, 36.4) (0.2, 34.3) (0.3, 32.6) (0.4, 27.0)
    (0.5, 23.5) (0.6, 22.1) (0.7, 21.4) (0.8, 22.1) (0.9, 24.2) (1.0, 28.7)
    };
\addplot[
    color=blue,
    mark=square,
    ]
    coordinates {
    (0, 4.1) (0.1, 5.1) (0.2, 7.0) (0.3, 7.0) (0.4, 10.6)
    (0.5, 15.1) (0.6, 20.8) (0.7, 26.0) (0.8, 52.0) (0.9, 54.5) (1.0, 55.6)
    };
\addplot[
    color=orange,
    mark=square,
    ]
    coordinates {
    (0, 26.5) (0.1, 34.2) (0.2, 35.6) (0.3, 35.6) (0.4, 40.0)
    (0.5, 44.6) (0.6, 54) (0.7, 59.6) (0.8, 68.9) (0.9, 67.4) (1.0, 67.4)
    };
\legend{Vicuna-13b, Llama-2-13b, Llama-2-70b}
\end{axis}
\end{tikzpicture}

\caption{Hybrid confidence elicitation}
\label{fig: hybrid_confidence_elicitation}
\end{figure}
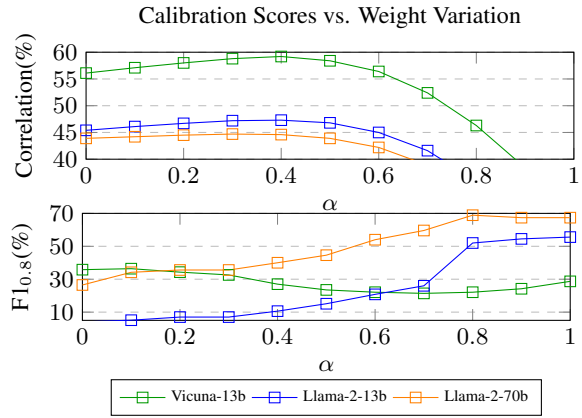

We explore whether combining self-evaluation and self-consistency can yield a more accurate confidence distribution on ASQA dataset. By blending confidence distributions from self-evaluation ($C^{eval}_i$) and self-consistency ($C^{consist}_i$) into a hybrid distribution $C^{hybrid}_i = \alpha C^{eval}_i + (1-\alpha)C^{consist}_i$, we adjust their relative contributions using $\alpha$. As shown in Figure \ref{fig: hybrid_confidence_elicitation}, we observe that the correlation between confidence and correctness initially increases but then declines as $\alpha$ varies from 0 to 1.
However, this trend doesn't extend to other metrics like F1, indicating that while hybrid calibration elicitation may enhance calibration in terms of correlation, it may not have the same impact on other dimensions.

\newpage
\section{Prompts}
This section introduces the prompts used for our experiments.
\subsection{Correctness Evaluation Template}
\label{appendix: correctness evaluation template}
Similar to contemporary work \citep{Kim2023WhichIB}, our evaluation template for GPT-4 evaluation to get the target correctness distribution of an answer includes four components: a clear task description, expertly crafted evaluation criteria for objectivity, demonstrations with a variety of answer qualities (best, worst, intermediate) each with a score and rationale, and specific evaluation instructions for the LLM, encompassing the question-answer pair to be evaluated and a reference answer.  
\begin{tcolorbox}[title=Evaluation Template]
\small
\{task instruction\}
\\ \\
You will be given a question, a reference answer, and a student's answer. Please evaluate the student's answer based on both your knowledge and the reference answer, and provide a score from 0-5 to the student's answer. Keep in mind that the reference answer is not the sole correct response. Assess for both factual accuracy and relevance to the question. The following are the scoring criterion:
\\ \\
\{criterion\}
\\ \\
Here are some examples.
\\ \\
\{examples\}
\\ \\
Now it's your turn.
\\ \\
Question: \{question\}
\\ \\
Reference answer: \{reference answer\}
\\ \\
Student's answer: \{answer\}
\\ \\
Now please provide your score about this answer in the format of ``Score: <Your score>/5'' and give your explanation.
\end{tcolorbox}

\subsection{Criteria}
Below are the criteria for various tasks, with a special note that the CNNDM summarization task utilizes a distinct evaluation template.
\label{appendix: criteria}
\\ \\
\textbf{ASQA Criterion}
\begin{tcolorbox}[title=ASQA Criterion]
\small
5 - Completely Correct and Highly Relevant: The answer fully addresses the question, resolves the ambiguity, and provides a well-rounded resolution. All facts presented in the answer are accurate and relevant.
\\
4 - Mostly Correct and Relevant: The answer is very relevant and addresses the ambiguity well, but might have a minor oversight or inaccuracy. All the facts presented are accurate and relevant, or with only minor errors.
\\
3 - Partially Correct and Relevant: The answer is generally on topic and attempts to address the ambiguity, but there might be inaccuracies or omissions. The majority of the facts are correct, with a few errors.
\\
2 - Flawed but Somewhat Relevant: The answer somewhat addresses the topic but does not fully explore the question's ambiguity or does not provide a complete resolution. The facts presented are a mix of correct and incorrect information, with about half being accurate.
\\
1 - Mostly Incorrect or Mostly Irrelevant: The answer slightly touches upon the topic but misses the main point. The majority of the facts presented are incorrect, with only a small portion being accurate.
\\
0 - Completely Incorrect or Completely Irrelevant: The student's answer is completely off-topic, not related to the question at all, or contains only incorrect information.

\end{tcolorbox}

\textbf{ELI5 Criterion}
\small
\begin{tcolorbox}[title=ELI5 Criterion]
5 - Perfectly Addressed, Accurate and Clarity: The answer flawlessly addresses the question with exceptional accuracy and clarity. It simplifies complex concepts effectively and does so in a way that is captivating and memorable.
\\
4 - Accurate and clear: The answer is accurate, relevant to the question, and presented in a way that is engaging and understandable. It simplifies complex concepts effectively but may miss a small opportunity for further clarification or engagement.
\\
3 - Moderately Accurate and Understandable: The answer is mostly accurate and somewhat understandable. It addresses the question reasonably well but may lack detail or contain some inaccuracies. It may use complex terms or concepts that are not broken down into simpler ideas. 
\\
2 - Relevant but Lacks Clarity or Accuracy: The answer is related to the question but lacks clarity or contains partial inaccuracies. It attempts to simplify the idea but does not do so effectively, leaving room for confusion or misunderstanding.
\\
1 - Significantly Flawed: The answer addresses the question to a minimal extent but contains significant inaccuracies or misleading information. It might show a basic attempt to simplify the concept but fail in accuracy or relevance.
\\
0 - Completely Inaccurate or Irrelevant: The answer is entirely off-topic, irrelevant, or factually incorrect. It fails to address the question and does not simplify complex ideas.
\end{tcolorbox}

\begin{tcolorbox}[title=QAMPARI Criterion]
\small
*   Estimated Accuracy Assessment (0-3 Points)
\\
    * 3 Points: All answers provided seem correct based on available knowledge.
    \\
    * 2 Points: Majority of the answers seem correct.
    \\
    * 1 Point: Some answers are correct, but there are notable inaccuracies.
    \\
    * 0 Points: No answers seem correct or very high degree of inaccuracy.
    \\ \\
* 		Estimated Completeness Assessment (0-2 Points)
\\
    * 2 Points: The response seems comprehensive, covering a broad range of known or expected correct answers.
    \\
    * 1 Point: The response covers some correct answers but misses significant known or expected answers.
    \\
    * 0 Points: The response is highly incomplete, missing most of the known or expected correct answers.
    \\ \\
* 		Total Score (0-5 Scale)
\\
    * Add the points from Estimated Accuracy and Estimated Completeness.

\end{tcolorbox}

\begin{tcolorbox}[title=Summarization Evaluation Template]
\small
You will be given one summary written for a news article.
\\ \\
Your task is to rate the overall quality of the summary with a score from 0 to 5, where 0 is the lowest and 5 is the highest.
\\ \\
Please make sure you read and understand these instructions carefully. Please keep this document open while reviewing, and refer to it as needed.
\\ \\
Evaluation Steps:
\\
1. Read the news article carefully and identify the main topic and key points.
\\
2. Read the summary and compare it to the news article. Check if the summary covers the main topic and key points of the news article, and if it presents them in a clear and logical order.
\\
3. Assign a score for the summary quality on a scale of 0 to 5, where 0 is the lowest and 5 is the highest.
\\ \\
Task Input:
\\ \\
Article:
\{article\}
\\ \\
Summary:
\{summary\}
\\ \\
Now please provide your score of the summary in the format of ``Score: <Your score>/5'' and give your explanation.
\end{tcolorbox}

\subsection{Self-Evaluation Template}
\small
\label{subappendix: self-evaluation template}
\begin{tcolorbox}[title=Self Evaluation Template]
\{task instruction\}
\\ \\
You will be given a question and a student's answer. Please evaluate the student's answer, and provide a score from 0-5 to the student's answer based on the following scoring criterion:
\\ \\
\{criterion\}
\\ \\
Here are some examples.
\\ \\
\{examples\}
\\ \\
Now it's your turn.
\\ \\
Question: \{question\}
\\ \\
Answer: \{answer\}
\\ \\
Now please provide your score about this answer in the format of ``Score: <Your score>/5'' and give your explanation. Assess for both factual accuracy and relevance to the question.
\end{tcolorbox}

\subsection{Self-Consistency Template}
\label{subsec: self-consistency template}
In self-consistency approaches, assessing the similarity between two answers requires the use of an additional LLM in naive and split/claim methods. For the self-consistency-naive method, this involves an LLM directly providing similarity scores for the pair of answers. Meanwhile, for the self-consistency-split/claim method, another LLM is tasked with detecting the presence of sentences from answer 1 in answer 2. The templates for both the naive method and the split/claim method are outlined below.
\begin{tcolorbox}[title=Similarity Comparison Naive Template]
\small
You will be presented with a question followed by two answers. Evaluate how similar these answers are, considering their amount of information provided, factual content, effectiveness in addressing the question, format and organization. Conclude by providing an integer similarity score between 0 and 5.

Question: \{question\}

Answer 1: \{answer1\}

Answer 2: \{answer2\}

Now please give your similarity score in the format of ``Similarity score: <Your score>/5'' and give your explanation. Make your answer short and concise.
\end{tcolorbox}

\begin{tcolorbox} [title=Similarity Comparison Split Template]
\small
You will receive a sentence and a response; please ascertain if a similar statement is present in the response.

Sentence: \{sentence\}

Response: \{response\}

Verify if the response contains a statement resembling the target sentence. Answer yes or no.
\end{tcolorbox}

\section{Qualitative Analysis}
We showcase successful and failed calibration examples on Qampari and ASQA with self-consistency method.

\begin{table*}[htbp]\small
    \centering
    {
      \begin{threeparttable}
        \begin{tabularx}{\textwidth}{M{3cm}P{12cm}}
          \toprule
            \textbf{type} & \textbf{content} \\
            \midrule
            \multicolumn{2}{c}{\cellcolor{gray!15}\textbf{QAMPARI Successful Example 1}} \\
            \midrule
            \textit{Question} & Who directed a film that had P. Balachandran as a screenwriter? \\
            \textit{Answer} & Kamal, P. Balachandran, T. K. Rajeev Kumar, V. K. Prakash, Rajeev Ravi.\\
            \textit{Model Answer} & Adu, Ammayum Muyalum, Bhoothakkannadi, Chattambi,  Kakkakku Pinnaka...\\
            \midrule 
            \textit{Other Answer 1} & Agraharathile Kazhuthai, Kaliyugam, Kodungallur, Kodiyettam ...\\ 
            \textit{Other Answer 2} &  Kodiyettam, Kuttyedam, Oru CBI Diary Kurippu...\\
            \textit{Other Answer 3} &  Aravindan, K.G. George, John Abraham, Shyam Benegal ... \\
            \textit{...} & ... \\
            \textit{Self-consistency scores} & [0, 0, 0, ...] (\%)\\
            \midrule
            \textit{Correctness} & Score: \makebox[1cm][l]{$0\%$} Distribution: $\{0 : 1.0\}$ \\
            \textit{Confidence score} & Score: \makebox[1cm][l]{$0\%$}  Distribution: $\{0: 1.0\}$\\
            \midrule

            \multicolumn{2}{c}{\cellcolor{gray!15}\textbf{QAMPARI Successful Example 2}} \\
            \midrule
            \textit{Question} & Which cities or municipalities are located in the province of Basilan? \\
            \textit{Answer} & Akbar, Al-Barka, Hadji Muhtamad, Isabela City, Lamitan, Lantawan, Maluso, Sumisip, Tabuan-Lasa, Tipo-Tipo, Tuburan, Ungkaya Pukan.\\
            \textit{Model Answer} & Al-Barka, Basilan, Isabela, Lamitan, Tipo-Tipo, Maluso, Sumisip.\\
            \midrule 
            \textit{Other Answer 1} & Basilan, Isabela, Lamitan, Lantawan, Tipo-Tipo, Sumisip.\\ 
            \textit{Other Answer 2} & Basilan, Tipo-Tipo, Isabela City, Maluso, Sumisip, Tuburan.\\
            \textit{Other Answer 3} & Basilan, Isabela, Tipo-Tipo, Lamitan, Sumisip, Al-Barka, Maluso, Tuburan. \\
            \textit{...} & ... \\
            \textit{Self-consistency scores} & [71.4, 57.1, 85.7, ...](\%)\\
            \midrule
            \textit{Correctness} & Score: \makebox[1cm][l]{$83.3\%$} Distribution: $\{0.8 : 1.0\}$ \\
            \textit{Confidence} & Score: \makebox[1cm][l]{$76.2\%$} Distribution: $\{0.6: 0.11, 0.8: 0.78, 1.0: 0.11\}$ \\
            \midrule
            
            \multicolumn{2}{c}{\cellcolor{gray!15}\textbf{QAMPARI Failed Example 1 (Under Confident)}} \\
            \midrule
            \textit{Question} & What city or municipality located in the Philipines are in the Tarlac province? \\
            \textit{Answer} & Anao, Bamban, Camiling, Capas, Concepcion, Gerona, La Paz, Mayantoc, Moncada, Paniqui, Pura, Ramos, San Clemente, San Jose, San Manuel, Santa Ignacia, Tarlac City, Victoria. \\
            \textit{Model Answer} & Capas, Concepcion, Gerona, La Paz, Mayantoc, Paniqui, San Jose, Santa Ignacia, Tarlac City, Victoria.\\
            \midrule 
            \textit{Other Answer 1} & Capas, Concepcion, Camiling, Tarlac City, Paniqui, San Jose, Santa Lucia, Victoria.\\ 
            \textit{Other Answer 2} & Capas, Tarlac, Concepcion, Tarlac, Gerona, Tarlac, La Paz, Tarlac, Paniqui, Tarlac, San Jose, Tarlac, Tarlac City.\\
            \textit{Other Answer 3} & Capas, Concepcion, Camiling, Tarlac City, San Jose. \\
            \textit{...} & ... \\
            \textit{Self-consistency scores} & [60, 60, 30, ...](\%)\\
            \midrule
            \textit{Correctness} & Score: \makebox[1cm][l]{$100\%$} Distribution: $\{1 : 1.0\}$ \\
            \textit{Confidence} & Score: \makebox[1cm][l]{$51.1\%$} Distribution: $\{0.4 : 0.56, 0.6: 0.33, 0.8: 0.11\}$ \\
            \midrule

            \multicolumn{2}{c}{\cellcolor{gray!15}\textbf{QAMPARI Failed Example 1 (Over Confident)}} \\
            \midrule
            \textit{Question} & What are some notables mailing list softwares ? \\
            \textit{Answer} & Dada Mail, Discourse, GNU Mailman, GroupServer, LISTSERV, Majordomo, Organic Groups Mailing List, Sympa.\\
            \textit{Model Answer} & Mailchimp, Constant Contact, Campaign Monitor, AWeber, GetResponse, MailPoet, Sendinblue, MailerLite, Klaviyo, ConvertKit.\\
            \midrule 
            \textit{Other Answer 1} & Mailchimp, AWeber, GetResponse, Campaign Monitor, Constant Contact, Klaviyo, ConvertKit, MailPoet, Mailster, Sendinblue, MailerLite.\\ 
            \textit{Other Answer 2} & Mailchimp, Constant Contact, AWeber, Campaign Monitor, GetResponse, Klaviyo, MailPoet, Mailster, Sendinblue, MailerLite, ConvertKit, ActiveCampaign, and Constant Contact.\\
            \textit{Other Answer 3} & Mailchimp, AWeber, GetResponse, Campaign Monitor, MailerLite, Constant Contact, Klaviyo, ConvertKit, Sendinblue, and Emma. \\
            \textit{...} & ... \\
            \textit{Self-consistency scores} & [100, 100, 90, ...](\%)\\
            \midrule
            \textit{Correctness} & Score: \makebox[1cm][l]{$0\%$} Distribution: $\{0 : 1.0\}$ \\
            \textit{Confidence score} & Score: \makebox[1cm][l]{$97.8\%$}  Distribution: $\{0.8: 0.22, 1.0: 0.78\}$\\
          \bottomrule
        \end{tabularx}
      \end{threeparttable}
    }
    \caption{\label{table:qualitative examples} Examples for QAMPARI}
\end{table*}

\begin{table*}[htbp]\small
    \centering
    {
      \begin{threeparttable}
        \begin{tabularx}{\textwidth}{M{3cm}P{12cm}}
          \toprule
            \textbf{type} & \textbf{content} \\
            \midrule
            \multicolumn{2}{c}{\cellcolor{gray!15}\textbf{ASQA Successful Example 1}} \\
            \midrule
            \textit{Question} & What are the three types of enzymes in digestion? \\
            \textit{Answer} & Digestive enzymes are a group of enzymes that break down polymeric... Lipases split fatty acids... Proteases and peptidases split proteins ... Amylases split carbohydrates ... Nucleases split nucleic acids into nucleotides...\\
            \textit{Model Answer} & There are three main types of enzymes involved in digestion: proteases, which break down proteins; amylases, which break down carbohydrates; and lipases, which break down fats. Proteases are produced by the stomach and pancreas, while amylases are ... These enzymes work in concert ...\\
            \midrule 
            \textit{Other Answer 1} & Digestive enzymes are proteins that help ... There are three main types of digestive enzymes: proteases, which break down proteins into amino acids; amylases, which break down carbohydrates into simple sugars; and lipases, which break down fats into glycerol and fatty acids. These enzymes are produced by different organs and ...\\ 
            \textit{Other Answer 2} & There are three main types of enzymes involved in digestion: proteases, which break down proteins; amylases, which break down carbohydrates; and lipases, which break down fats. These enzymes are produced by the salivary glands, stomach, and pancreas, and work in various parts of the digestive system...\\
            \textit{...} & ... \\
            \textit{Self-consistency scores} & [100, 100, 0, ...] (\%)\\
            \midrule
            \textit{Correctness} & Score: \makebox[1cm][l]{$100\%$} Distribution: $\{1.0 : 1.0\}$ \\
            \textit{Confidence score} & Score: \makebox[1cm][l]{$95.6\%$}  Distribution: $\{0.8: 0.22, 1.0: 0.78\}$\\
            \midrule

            \multicolumn{2}{c}{\cellcolor{gray!15}\textbf{ASQA Successful Example 2}} \\
            \midrule
            \textit{Question} & When did ben rectors new album come out?\\
            \textit{Answer} & Benjamin Evans Rector is an American singer... He has released seven studio albums, including Magic on June 22, 2018, which peaked at No. 1 .. Brand New which released on August 28, 2015... Older albums such as Something Like This, released on September 13, 2011 and The Walking in Between, came out August 20, 2013.\\
            \textit{Model Answer} & Ben Rector's new album, "Magic", was released on March 25, 2022.\\
            \midrule 
            \textit{Other Answer 1} & Ben Rector's new album, "Brand New", was released on March 10th, 2022.\\ 
            \textit{Other Answer 2} & Ben Rector, an American singer-songwriter, released his new album "Magic" on October 20...\\
            \textit{...} & ... \\
            \textit{Self-consistency scores} & [0, 0, ...](\%)\\
            \midrule
            \textit{Correctness} & Score: \makebox[1cm][l]{$0\%$} Distribution: $\{0 : 1.0\}$ \\
            \textit{Confidence} & Score: \makebox[1cm][l]{$0\%$} Distribution: $\{0: 1.0\}$ \\
            \midrule
            
            \multicolumn{2}{c}{\cellcolor{gray!15}\textbf{ASQA Failed Example 1 (Under Confident)}} \\
            \midrule
            \textit{Question} & Who plays charlie on it's always sunny? \\
            \textit{Answer} & Charlie Day plays the fictional character Charlie Kelly... Day is an American actor, screenwriter... \\
            \textit{Model Answer} & Charlie Day plays Charlie on It's Always Sunny in Philadelphia. He is also known for his role as Dale Arbus ... Day is a member of the improvisational comedy ...\\
            \midrule 
            \textit{Other Answer 1} & Dennis is played by Glenn Howerton.\\ 
            \textit{Other Answer 2} & Charlie on It's Always Sunny in Philadelphia is played by Charlie Day.\\
            \textit{...} & ... \\
            \textit{Self-consistency scores} & [0, 33.3, ...](\%)\\
            \midrule
            \textit{Correctness} & Score: \makebox[1cm][l]{$80\%$} Distribution: $\{0.8 : 1.0\}$ \\
            \textit{Confidence} & Score: \makebox[1cm][l]{$31.1\%$} Distribution: $\{0 : 0.22, 0.4: 0.78\}$ \\
            \midrule

            \multicolumn{2}{c}{\cellcolor{gray!15}\textbf{ASQA Failed Example 2 (Over Confident)}} \\
            \midrule
            \textit{Question} & Who sings wake me up when it's all over?\\
            \textit{Answer} & 
            ``Wake Me Up'' is a song by Swedish DJ and record producer Avicii. Aloe Blacc, which is the stage name for Egbert Nathaniel Dawkins III provides vocals for the song while Mike Einziger provides acoustic guitar. ``Wake Me When It's Over'' is a song written and recorded by Willie... \\
            \textit{Model Answer} & ``Wake Me Up When September Ends'' is a song written by Billy\\
            \midrule 
            \textit{Other Answer 1} & ``Wake Me Up When September Ends'' is a song by American rock band Green Day, written by lead vocalist Billie Joe Armstrong...\\ 
            \textit{Other Answer 2} & ``Wake Me Up When September Ends'' is a song by Green Day, written by the band's lead vocalist and guitarist Billie Joe Armstrong...\\
            \textit{...} & ... \\
            \textit{Self-consistency scores} & [100, 100, ...](\%)\\
            \midrule
            \textit{Correctness} & Score: \makebox[1cm][l]{$0\%$} Distribution: $\{0 : 1.0\}$ \\
            \textit{Confidence score} & Score: \makebox[1cm][l]{$77.8\%$}  Distribution: $\{0: 0.22, 1.0: 0.78\}$\\
          \bottomrule
        \end{tabularx}
      \end{threeparttable}
    }
    \caption{\label{table:qualitative examples asqa} Examples for ASQA}
\end{table*}

\end{document}